%% file: main.tex
\documentclass[10pt]{article} 
\usepackage[preprint]{tmlr}

\input{math_commands.tex}

\usepackage{hyperref}
\usepackage{url}
\input{packages}

\title{CP-JDOT: Conformal Risk Minimization for Semi-Supervised Domain Adaptation via Optimal Transport}

\author{\name Manos Giannopoulos \email manos.giannopoulos@duke.edu \\
      \addr Duke University
      \AND
      \name Yi Shen \email ys267@duke.edu \\
      \addr Duke University
      \AND
      \name Michael M. Zavlanos \email mz61@duke.edu \\
      \addr Duke University
      }

\def\month{08}  
\def\year{2026} 
\def\openreview{\url{https://openreview.net/forum?id=XXXX}} 

\begin{document}

\maketitle

\begin{abstract}
In high-stakes healthcare applications, machine learning models are frequently trained on data from one patient population and deployed on another, creating a distribution shift that degrades their performance. Semi-Supervised Domain Adaptation (SSDA) addresses this by leveraging labeled data from some source domain to improve model performance on a target domain where labels are scarce. However, existing SSDA methods optimize primarily for point-prediction accuracy and offer no principled uncertainty quantification --- a prerequisite for clinical trust. Conformal Prediction (CP) can address this limitation by providing prediction sets with rigorous, distribution-free coverage guarantees. However, applying CP post-hoc to a pre-trained model can yield prohibitively large prediction sets, as SSDA pre-training methods do not account for the nonconformity score geometry that determines conformal set size. Conformal Risk Minimization (CRM) has been used to resolve this issue in the fully supervised setting by integrating the CP objective directly into model training, but it requires a large labeled dataset to compute nonconformity thresholds during training, precisely the data that is scarce in the SSDA regime. We propose CP-JDOT, an end-to-end framework that integrates CRM into the SSDA training objective, enabling effective CRM in the limited-labeled-target-data regime. The key idea is to utilize Optimal Transport (OT) to generate pseudolabels for unlabeled target instances, providing the additional training signal needed by CRM to operate using only a small labeled target set. This results in a model jointly optimized for domain alignment and conformal set size, producing compact prediction sets with marginal coverage guarantees and allowing us to incorporate domain knowledge into training, as we show by penalizing the inclusion of both benign and malignant classes in a skin lesion classification task.\footnotemark
\end{abstract}
\footnotetext{Large Language Models (ChatGPT and Claude) were used
to assist with spell checking, grammar, wording, and writing code
to produce the figures.}

\section{Introduction}

\input{sections/introduction}

\section{Preliminaries} \label{section:prelim}

\input{sections/preliminaries}

\section{Related Work}

\input{sections/relatedwork}

\section{Problem Definition} \label{section:problem}

\input{sections/definition}

\section{Method} \label{section:method}

\input{sections/method}

\section{Experiments} \label{section:results}

\input{sections/results} 

\section{Conclusion}

\input{sections/conclusion}

\FloatBarrier
\bibliography{main}
\bibliographystyle{tmlr}

\newpage

\appendix
\section*{Appendix}

\section{Geometric Optimal Transport summary} \label{app:got}

\input{appendix/got}

\section{Experimental Details} \label{app:details}

\input{appendix/experimental_details}

\section{Additional Experimental Results} \label{app:additional}

\input{appendix/extended_results}

\end{document}

%% file: math_commands.tex
\usepackage{amsmath,amsfonts,bm}

\def\eqref#1{equation~\ref{#1}}

\def\1{\bm{1}}

\DeclareMathAlphabet{\mathsfit}{\encodingdefault}{\sfdefault}{m}{sl}
\SetMathAlphabet{\mathsfit}{bold}{\encodingdefault}{\sfdefault}{bx}{n}



%% file: packages.tex
\usepackage{mathtools}
\usepackage{amsthm}
\usepackage{appendix}
\usepackage{yhmath}
\usepackage{algorithm}
\usepackage{algpseudocode}
\usepackage{tikz}
\usetikzlibrary{shapes.geometric, arrows}
\usepackage{amsfonts}
\usepackage{stackengine}
\usepackage[margin=1in]{geometry}
\usepackage{amssymb, amsmath}
\usepackage{hyperref}
\usepackage{float}
\usepackage{pdfpages}
\usepackage{graphicx}
\usepackage{bbm}
\usepackage{enumitem} 
\usepackage{subcaption}
\usepackage{booktabs}
\usepackage{placeins}

\DeclarePairedDelimiterX{\norm}[1]{\lVert}{\rVert}{#1}

\tikzstyle{phase} = [rectangle, rounded corners, minimum width=6cm, minimum height=1.5cm, text centered, draw=black]
\tikzstyle{arrow} = [thick,->,>=stealth]

%% file: sections/introduction.tex
In many clinical applications of machine learning, medical imaging being a prominent example, models trained in one setting can perform worse when applied to another \citep{zech2018generalization,daneshjou2022disparities}. In these settings, labeled data in the target domain are often scarce because obtaining expert annotations is costly and time-consuming \citep{willemink2020preparing}, while labeled data from a related source domain and unlabeled data from the target domain may be abundant. Semi-Supervised Domain Adaptation (SSDA) algorithms exploit this setting by leveraging labeled source data and unlabeled target data alongside the limited labeled target data to improve point-prediction accuracy on the target domain. In high-stakes settings, where an incorrect decision could have serious consequences, a clinician may also need a calibrated measure of uncertainty to assess when a prediction can be trusted and when further evidence is warranted. Conventional SSDA objectives focus on point-prediction accuracy, leaving this uncertainty to be quantified separately.
 
\textit{Conformal Prediction} \citep{Vovk2005} (CP) can address this limitation by equipping any pre-trained model with distribution-free, finite-sample guarantees on prediction uncertainty. Given a significance level \( \alpha \), CP constructs a prediction set \( C(X) \) for an input \( X \) (e.g., a feature vector, an image, or other structured input) such that the true label $Y$ is included in the set with a probability of at least $1 - \alpha$, i.e.,
\[
\mathbb{P}(Y \in C(X)) \geq 1 - \alpha.
\]
Through this guarantee, CP produces prediction sets that are inherently adaptive: small for inputs that the model can easily classify, and larger for inputs that are more difficult or atypical. This adaptive behavior makes predictive uncertainty explicit in a statistically rigorous way, providing a principled foundation for clinical decision-making.
 
Traditionally, CP has been applied as a post-hoc wrapper on top of pre-trained models \citep{AngelopoulosICLR2020,RomanoNIPS2019,RomanoNIPS2020,VovkCCPPA2018,VovkCOPA2020}. However, this approach often yields prediction sets that are far too large to be clinically useful. The reason is an objective mismatch: A pre-training objective that optimizes for point-prediction accuracy does not account for the nonconformity score geometry that governs conformal set size. While an optimized post-hoc calibration method can yield modest improvements in set size, the best achievable result is fundamentally constrained by the predictive outputs of the underlying model. Conformal Risk Minimization (CRM) \citep{BellottiCOPA2020,BellottiARXIV2021,stutz2021learning} addresses this limitation by incorporating set-size directly into the training objective via differentiable nonconformity thresholds, optimizing for conformal efficiency rather than top-1 accuracy. However, extending CRM to an SSDA setting is non-trivial: existing approaches rely on large labeled datasets to estimate nonconformity score thresholds during training, and do not provide a clear way to use labeled data from a different distribution. The question we address in this work is therefore:
 
\begin{center}
    \emph{How can we optimally train conformal predictors for semi-supervised domain adaptation tasks\\ where labeled data is limited in the target domain?}
\end{center}

\begin{figure}[t]
    \centering
    \includegraphics[width=\linewidth]{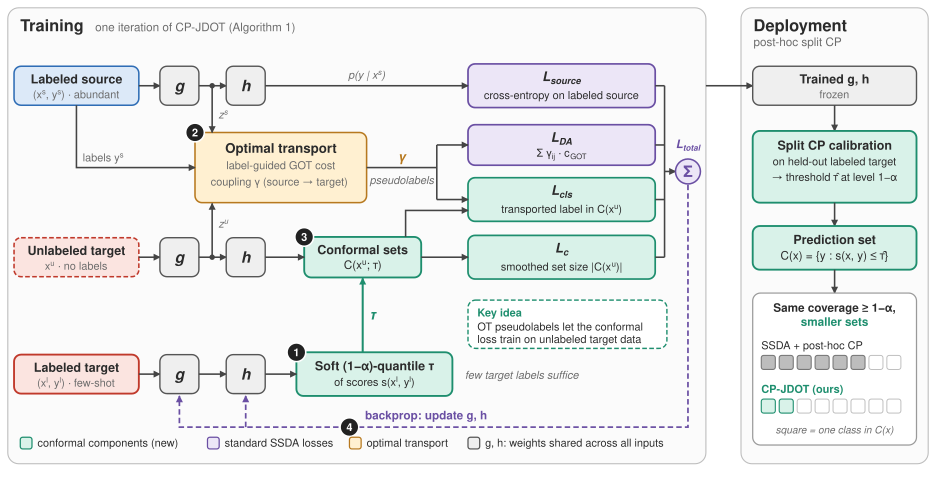}
    \caption{Our proposed framework, CP-JDOT, integrates in-the-loop conformal optimization for semi-supervised domain adaptation. Our algorithm uses the few labeled target samples to compute a differentiable quantile of the nonconformity score distribution. It uses this threshold both to solve the optimal transport problem and to construct conformal sets on the unlabeled target samples. The training objective includes a domain alignment component and a conformal component that aims to minimize set sizes while encouraging the inclusion of pseudolabels obtained by transporting source label information to the target domain. After training, we calibrate the model on a separate held-out set of labeled target samples.}
    \label{fig:conformal-method}
\end{figure}

In this work, we propose CP-JDOT (outlined in Figure \ref{fig:conformal-method}), an end-to-end CRM training algorithm designed for SSDA tasks where labeled target data are scarce. We assume access to a small labeled target set, sufficient unlabeled target data, and labeled source data. The labeled target samples are used to compute a differentiable quantile of the nonconformity scores at each training iteration, which serves as the conformal threshold during training. To align source and target distributions, we use Joint Domain Adaptation to map both domains into a shared representation space, and rely on Optimal Transport (OT) to transfer source labels to unlabeled target instances as pseudolabels. Conformal sets are then constructed around these pseudolabeled target instances using the quantile derived from the small labeled target set, and the resulting classification and set-size losses are backpropagated end-to-end, directly shaping the model's output geometry for conformal efficiency. Our experiments show that CP-JDOT consistently reduces average prediction set size compared to state-of-the-art SSDA pre-training methods that optimize for top-1 accuracy, while preserving coverage. Furthermore, since our method directly modifies the training loss, it can support domain-specific constraints --- for example, penalizing sets that simultaneously include benign and malignant classes in skin lesion classification --- making the framework attractive for applications where prediction set composition directly impacts clinical decision-making.
 
The key contributions of this work are summarized as follows:
 
\begin{enumerate}
 
\item We propose CP-JDOT, a novel SSDA training framework that integrates a CRM objective within a Joint Domain Adaptation with Optimal Transport framework to enable the optimization of a predictive model with respect to the prediction sets it will produce downstream. This framework specifically addresses target-label scarcity by leveraging OT-derived pseudolabels to supervise the conformal training loss while aligning the source and target latent representations, bridging the gap between supervised CRM and SSDA.
 
\item We demonstrate that, by incorporating the conformal objective directly into the training loop, CP-JDOT supports penalty terms tailored to domain-specific requirements. Beyond minimizing set cardinality, this enables control over set composition --- such as mitigating ``coverage confusion'' \citep{stutz2021learning} by penalizing the simultaneous inclusion of mutually exclusive high-level classes --- yielding prediction sets that are both statistically valid and semantically meaningful.
 
\item We provide a rigorous empirical analysis demonstrating that our end-to-end approach consistently outperforms state-of-the-art CP-agnostic SSDA methods with post-hoc conformal wrappers when evaluated on the downstream average set size, while preserving coverage guarantees.
 
\end{enumerate}

%% file: sections/preliminaries.tex
We will first review some key concepts around CP and CRM. Consider a supervised classification problem with $K$ classes, where inputs lie in a feature space $\mathcal{X}$ and labels in a label space $\mathcal{Y} = \{1, \dots, K\}$. Let $\mathcal{D}$ denote the joint distribution over $\mathcal{X} \times \mathcal{Y}$, and let $f_\theta$ be a probabilistic classifier modeled by a neural network parameterized by $\theta$. Moreover, let $f_{\theta,y}(x)$ denote the predicted probability of class $y \in \mathcal{Y}$ for a single input $x \in \mathcal{X}$. In CP, a pre-trained $f_\theta$ is usually ``wrapped'' to produce prediction sets $C_\theta(X)$, which are subsets of $\mathcal{Y}$ intended to contain the true label $Y$ with a desired probability, a constraint commonly called \emph{coverage}.

\subsection{Coverage Constraints in Conformal Prediction}

In CP, \emph{coverage} refers to the probability that the prediction set $C_{\theta}(X)$ contains the true label $Y$. In an ideal scenario, one would aim for \emph{full conditional coverage}, which requires that for every possible input $x \in \mathcal{X}$, the prediction set captures the true label with a probability of at least $1-\alpha$:

\begin{equation}
    \mathbb{P}\big(Y \in C_{\theta}(X) \mid X = x\big) \geq 1 - \alpha, \quad \text{for all } x \in \mathcal{X}.
    \label{eq:full-conditional-coverage}
\end{equation}

Intuitively, this guarantees that the method is perfectly calibrated for every possible input. However, achieving full conditional coverage is generally impossible in a distribution-free, finite-sample setting unless the algorithm adopts the trivial behavior of including the entire label space in the conformal sets \citep{lei2014distribution}. To address this limitation, a more achievable relaxation is \emph{marginal coverage}, which only requires that the coverage guarantee holds on average over the distribution of the features $X$:

\begin{equation}
\mathbb{P}\big(Y \in C_{\theta}(X)\big) \geq 1 - \alpha.
\label{eq:marginal-coverage}
\end{equation}

While this guarantee is weaker, it is attainable in a finite-sample, distribution-free setting using standard conformal prediction methods. Marginal coverage ensures that the prediction sets are correct on average, yet it does not provide guarantees for specific regions of the input space, which may result in under-coverage for certain subpopulations.

\subsection{Conformal Predictors}

In this work, we focus on \emph{Split Conformal Prediction} (SCP)~\citep{papadopoulos2002splitcp}, a practical variant of conformal prediction that guarantees marginal coverage using a held-out calibration dataset that has not been used during the training of the predictive model $f_\theta$. Let $D_{\text{cal}} = \{(x_i, y_i)\}_{i=1}^{N_{\text{cal}}}$ denote this calibration dataset, and let $s: \mathcal{X} \times \mathcal{Y} \rightarrow \mathbb{R}$ denote a \emph{nonconformity score} function. Intuitively, $s(\cdot,\cdot)$ measures how \emph{nonconforming} a pair $(x,y)$ is to the underlying data distribution $\mathcal{D}$. SCP involves two phases: (i) a \emph{calibration} phase, where a threshold $\tau$ is computed using $D_{\text{cal}}$ to satisfy the desired coverage level, and (ii) a \emph{prediction} phase, where prediction sets $C_\theta(x;\tau)$ are formed for individual inputs $x$. These sets depend on the model parameters $\theta$ both through the predicted probabilities $f_{\theta,y}(x)$ and the calibrated threshold $\tau$. We next review two commonly used conformal predictors that will be the basis of our experimental evaluation: HPS and APS.

The \textbf{Homogeneous Prediction Sets (HPS)}~\citep{SadinleJASA2019} predictor constructs confidence sets by thresholding the predicted probabilities:
\begin{equation}
    C_\theta(x;\tau) := \{y : s_\theta(x,y) \le \tau\}.
\end{equation}
where $s_\theta(x,y) = 1 - f_{\theta,y}(x)$. The subscript in $C_\theta$ emphasizes the dependence on the model $f_\theta$ and its parameters $\theta$. During calibration, the threshold $\tau$ is set to the $(1-\alpha)(1 + 1/N_{\text{cal}})$ quantile of the nonconformity scores $s_\theta(x_i, y_i)$ computed on the calibration set. This ensures that marginal coverage of at least $1-\alpha$ is achieved for new test examples $(X,Y)$, since the held-out calibration data and the new test sample are exchangeable.

The \textbf{Adaptive Prediction Sets (APS)}~\citep{RomanoNIPS2020} predictor constructs confidence sets based on the cumulative sums of sorted class probabilities. Specifically, the set is defined as $C_\theta(x;\tau) := \{y : s_\theta(x,y) \le \tau\}$, where the nonconformity score $s_\theta(x,y)$ is given by
\begin{align*}
    s_\theta(x, y) := \sum_{j=1}^{k-1} f_{\theta,y^{(j)}}(x) + U \cdot f_{\theta,y^{(k)}}(x),
\end{align*}
with $f_{\theta,y^{(1)}}(x) \ge f_{\theta,y^{(2)}}(x) \ge \dots \ge f_{\theta,y^{(K)}}(x)$ the sorted class probabilities and $U \sim \text{Unif}(0,1)$ used to break ties between different class scores. Calibration proceeds as in HPS: the threshold $\tau$ is set to the $(1-\alpha)(1 + 1/N_{\text{cal}})$ empirical quantile of the scores at the true labels, ensuring marginal coverage due to the exchangeability between the calibration samples and the test sample.

In our experiments, SCP using the above conformal predictors will be applied post-hoc to pre-trained models, producing prediction sets. Coverage is guaranteed regardless of the pre-training method due to the post-hoc recalibration. Therefore, the standard comparison metric is the \textbf{inefficiency} or the \textbf{average set size} (we use these two terms interchangeably throughout the manuscript). For a test set $T$ of size $n$, inefficiency is defined as
\begin{align*}
    \text{Inefficiency} := \frac{1}{n} \sum_{i \in T} |C_\theta(x_i)|,
\end{align*}
where $| \cdot |$ denotes the cardinality of the prediction set.

\subsection{Conformal Risk Minimization} \label{section:crm}

CRM aims to train a classifier such that the prediction sets it produces after downstream conformal calibration have minimal cardinality. A common formulation augments the standard Empirical Risk Minimization (ERM) objective with a \emph{conformal alignment} loss:

\begin{align}
\min_{\theta} \quad \mathcal{L}_{cls}(f_\theta) + \lambda \mathcal{L}_c(f_\theta) \label{eqn:crm}
\end{align}

where $\mathcal{L}_{cls}$ is a classification loss, $\mathcal{L}_c(f_\theta)$ is the conformal alignment loss, and $\lambda$ controls the trade-off between the two objectives. As in previous work \citep{stutz2021learning, BellottiCOPA2020}, we consider the case where the conformal alignment loss is the expected (smoothed) set cardinality parameterized by a quantile $\tau$:

\begin{align*}
\mathcal{L}_c(f_\theta ; \tau) = \mathbb{E}\left[\sum_{y \in \mathcal{Y}}\tilde{\mathbbm{1}}[s(X,y) \leq \tau]\right].
\end{align*}

Here, $\tilde{\mathbbm{1}}[\cdot]$ is a smoothed approximation of the indicator function $\mathbbm{1}[\cdot]$, implemented using the sigmoid function. Standard stochastic CRM methods employ a two-stage process within each training iteration, since the quantile $\tau$ of the score distribution changes with each model update. First, they sample a batch of labeled examples to estimate the current conformal quantile $\tau$. Then, they sample a separate batch to compute the classification and conformal alignment losses in problem \ref{eqn:crm}. While the conformal alignment loss $\mathcal{L}_c$ can be evaluated without access to ground-truth labels, the same is not true for the classification loss $\mathcal{L}_{cls}$ and the nonconformity score quantile $\tau$. Therefore, the main challenge our method aims to address is how to optimally use the few labeled target samples together with the labeled source samples in order to construct a CRM objective while bridging the two domains in an SSDA setting. 

%% file: sections/relatedwork.tex
\subsection{Domain Adaptation}

Unsupervised domain adaptation (UDA) considers prediction in a target domain for which labeled examples are unavailable during training, while labeled data are provided from a related source domain~\citep{BenD07}. A central challenge is to extract information that transfers across the domain shift without discarding features that are predictive of the target task. Existing approaches address this challenge through several mechanisms, including feature-distribution matching~\citep{deepcoral,LongMMD}, adversarial domain alignment~\citep{Ganin2016,Tzeng15}, and optimal transport.

Optimal Transport (OT) provides a distribution-level framework for establishing correspondences between samples from two domains~\citep{Kantorovich42,Villani09}. In domain adaptation, this correspondence can be used to transfer information from labeled source observations to otherwise unlabeled target observations~\citep{Cou14}. Joint Distribution Optimal Transport (JDOT)~\citep{courty2017joint} makes this correspondence sensitive to the prediction task by incorporating label discrepancies into the transport cost, while DeepJDOT~\citep{damodaran2018deepjdot} integrates this formulation with deep representation learning. The resulting transport plan associates target samples with source labels and can therefore serve as a form of supervision for the target domain. DeepJDOT can be considered a seminal work in the field; subsequent work has built upon it to improve the scalability and robustness of OT-based adaptation by considering Unbalanced and Partial Optimal Transport~\citep{fatras2021uot, nguyen2022pot}, or utilizing alternative distance metrics to the Euclidean distance proposed by DeepJDOT~\citep{tai2021sinkhorn, duque2022diffusion, asadulaev2024neural, gu2026keypoint, maman2025tmlr}. Our work also builds upon the framework proposed by DeepJDOT; by incorporating a CRM term and appropriately modifying its training procedure, we align the training objective more closely with conformal set-size efficiency, resulting in more compact prediction sets downstream.

Semi-supervised domain adaptation (SSDA) relaxes the UDA setting by adding a small labeled subset of target-domain observations. Existing SSDA methods exploit these labels through mechanisms such as entropy minimization~\citep{saito2019semi}, adversarial pseudo-label refinement~\citep{li2021cdac}, active target-label selection~\citep{he2024eftl}, and prototype-based or co-training strategies~\citep{huang2023promm, yang2021decota}. Our work considers the same limited-target-label setting but is complementary to these approaches: rather than using the labeled target data solely for conventional adaptation, we use them to incorporate the conformal objective into training, while leveraging OT-based domain adaptation to provide supervision for the unlabeled target data. To the best of our knowledge, we are the first to incorporate CRM into the SSDA framework. We therefore compare our approach against post-hoc conformalized versions of representative SSDA methods to validate the benefit of incorporating the conformal objective into the training procedure when the metric of interest is the downstream conformal set size.

\subsection{Conformal Prediction}

Conformal prediction (CP), introduced by~\cite{Vovk2005}, provides finite-sample prediction sets with marginal coverage guarantees under exchangeability, with applications to both regression and classification problems~\citep{RomanoNIPS2019, RomanoNIPS2020}. Split conformal prediction, the method adopted in this paper, is particularly common in practice due to its simplicity and compatibility with arbitrary predictive models~\citep{papadopoulos2002splitcp, lei2014distribution}. Nevertheless, valid conformal methods based on leave-one-out estimators~\citep{BarberARXIV2019b} and cross-calibration~\citep{VovkAMAI2013} are also present in the literature. Modern CP research focuses on improving non-conformity scores targeting smaller conformal set sizes~\citep{SadinleJASA2019, AngelopoulosICLR2020, huang2024saps} and better conditional coverage~\citep{RomanoNIPS2020}. However, the above methods all work as post-hoc calibrators. Thus, their performance is inherently constrained by the quality of the underlying predictive model. Prior work~\citep{BellottiARXIV2021} has noted that this model dependency is consequential: models trained with objectives that are misaligned with conformal set-size efficiency, such as the standard categorical cross-entropy loss that is commonly used to maximize top-1 accuracy, may yield larger and less informative prediction sets after calibration. 

Conformal Risk Minimization (CRM), introduced concurrently by~\cite{BellottiARXIV2021} and~\cite{stutz2021learning}, aims to \emph{pre-train} a model to learn better non-conformity scores, leading to reduced downstream inefficiency (i.e., smaller set sizes) after the post-hoc calibration. Following the seminal paper of~\cite{stutz2021learning} (ConfTr), several works have proposed improved versions of CRM, including enhanced conditional coverage (CUT)~\citep{einbinder2022cut}, reduced variance (ConfTr-VR)~\citep{noorani2025}, robustness to adversarial samples (RSCP)~\citep{yan2024robustcp}, tighter learning bounds (DPSM)~\citep{shi2025dpsm}, and learning from long tailed data~\citep{marani2026classadaptiveconformaltraining}. The above approaches all consider supervised learning problems; our work extends the approach of CRM to SSDA, enabling the learning of non-conformity scores without access to a high volume of labeled data samples from the target distribution.

%% file: sections/definition.tex
We consider a \emph{multi-class classification} problem across a source and a target domain, characterized by a shared input space $\mathcal{X}$ and a common label space $\mathcal{Y}$, with $|\mathcal{Y}| = K$.  Let $\mathcal{S} = \{(x_i^s, y_i^s)\}_{i=1}^{N_s}$ denote a labeled source dataset, with samples drawn i.i.d. from a joint distribution $\mathcal{D}^S$ over $\mathcal{X} \times \mathcal{Y}$. The target domain is drawn from a different joint distribution $\mathcal{D}^T \neq \mathcal{D}^S$ over the same space and consists of a small labeled subset $\mathcal{T}_l = \{(x_j^l, y_j^l)\}_{j=1}^{N_l}$ and a larger unlabeled subset $\mathcal{T}_u = \{x_k^u\}_{k=1}^{N_u}$. 

Our goal is to solve a \emph{Conformal Risk Minimization} (CRM) problem, similar to the one described in Section \ref{section:crm}. We model the predictive function $f_\theta: \mathcal{X} \rightarrow \Delta(\mathcal{Y})$ as a neural network with parameters $\theta$, where $\Delta(\mathcal{Y})$ denotes the probability simplex over the label space $\mathcal{Y}$. At deployment, the trained model is wrapped by a post-hoc \emph{conformalizing} function $\phi: \Delta(\mathcal{Y}) \rightarrow 2^\mathcal{Y}$ to produce prediction sets $C_\theta \coloneq \phi \circ f_\theta$. Our goal is to learn $f_\theta$ such that the resulting conformal sets have small expected cardinality on the target domain while maintaining the desired coverage level. As mentioned previously, since the post-hoc conformalization procedure provides the desired coverage guarantee, the role of CRM during training is to improve the efficiency of the resulting prediction sets.

In addition to conformal efficiency, the model must adapt to the target domain using the limited labeled target data together with the available source and unlabeled target data. We therefore consider objectives that combine a classification loss, a domain adaptation loss, and a conformal alignment loss. Specifically, we define a \emph{Domain Adaptation Conformal Risk Minimization} (DA-CRM) problem as an objective of the form

\begin{align} \label{problem:ssda-crm}
\min_{\theta} \quad \mathcal{L}_{cls}(f_{\theta}) + \lambda_1 \mathcal{L}_{DA}(f_{\theta}) + \lambda_2 \mathcal{L}_{c}(f_{\theta}).
\end{align}

In Problem \ref{problem:ssda-crm}, $\mathcal{L}_{cls}$ promotes predictive accuracy, $\mathcal{L}_{DA}$ encourages adaptation between the source and target domains, and $\mathcal{L}_{c}$ acts as the regularization term that leads to conformal efficiency by reducing the size of the resulting prediction sets. The coefficients $\lambda_1$ and $\lambda_2$ control the relative contributions of the domain adaptation and conformal objectives. The following section describes our specific choices for these loss functions and the resulting training algorithm.

%% file: sections/method.tex
In this section, we present our method for constructing and optimizing Problem \ref{problem:ssda-crm}. Before presenting our method, we will first review DeepJDOT, an established Optimal Transport-based domain adaptation framework that forms the basis of our approach. We then introduce CP-JDOT and describe how we extend DeepJDOT to incorporate CRM in its training objective.

\subsection{Joint Domain Adaptation with Optimal Transport: The DeepJDOT framework}

In order to adapt CRM to an SSDA setting, we build upon the framework proposed in DeepJDOT \citep{damodaran2018deepjdot}. DeepJDOT decomposes $f_{\theta}$ into two distinct components: an embedding function $g: \mathcal{X} \rightarrow \mathcal{Z}$ that maps an input from either the source or target domain to a latent space $\mathcal{Z}$, and a classifier $h: \mathcal{Z} \rightarrow \mathcal{Y}$ that maps these latent representations to the label space. The goal is to jointly optimize the feature representation and classifier so that the learned representation captures features that are useful for classification while aligning the source and target domains. To this end, they define the following transport optimization problem:

\begin{equation}
\min_{\gamma\in \Pi(\mu_s,\mu_t),\, h,g}\quad 
\sum_i\sum_j \gamma_{ij} 
\, d \Big( g(x_i^s), y_i^s;\,
g(x_j^u), h(g(x_j^u)) \Big),
\label{eq:jdot} 
\end{equation}
where 
\begin{equation}
d \Big( g(x_i^s), y_i^s;\,
g(x_j^u), h(g(x_j^u)) \Big)
=
\lambda_a \, c\big(g(x_i^s), g(x_j^u)\big)
+
\lambda_{cls} \, \mathcal{L}_{cls}\big(y_i^s, h(g(x_j^u))\big).
\end{equation}

Here, $\Pi(\mu_s, \mu_t)$ denotes the set of all joint distributions with marginals $\mu_s$ and $\mu_t$, $c(\cdot, \cdot)$ denotes an appropriate cost function that measures the cost of transporting $g(x_i^s)$ to $g(x_j^u)$, and $\mathcal{L}_{cls}(\cdot, \cdot)$ denotes a classification loss. The parameters $\lambda_a$ and $\lambda_{cls}$ control the tradeoff between the alignment and classification terms. The first term in $d$ encourages alignment between the source and target distributions by minimizing a distance metric between their latent representations. The second term evaluates the classifier $h$ on the target domain by measuring its agreement with the labels of the source samples under the optimal transport plan. In the next section, we show how we define $c(\cdot,\cdot)$ and $\mathcal{L}_{cls}(\cdot,\cdot)$ in CP-JDOT so that the resulting objective implements CRM, explicitly accounting for both representation geometry and conformal set efficiency.

\subsection{Incorporating CRM with CP-JDOT}

We restate our method's end goal: to learn a model $f_\theta$ which, when paired with a downstream conformal prediction wrapper, produces more efficient prediction sets than a standard SSDA-trained model with post-hoc calibration. To achieve this, CP-JDOT incorporates conformal prediction into the training procedure while jointly optimizing the embedding function $g$ and classifier $h$.

At a high level, each training iteration consists of three steps. First, we estimate a differentiable $(1-\alpha)$-quantile $\tau$ of the nonconformity scores using the labeled target set $\mathcal{T}_\ell$. Second, we use labeled source and unlabeled target samples to compute an Optimal Transport coupling, which aligns the two domains and provides source-label information for the target samples. Finally, we use this coupling together with $\tau$ to optimize a combined objective that encourages both domain alignment and conformal efficiency.

We implement this procedure stochastically using two minibatches of size $B$ at each iteration: $\mathcal{B}_s = \{(x_i^s, y_i^s)\}_{i=1}^B$ sampled from $\mathcal{S}$ and $\mathcal{B}_t = \{x_j^{u}\}_{j=1}^B$ sampled from $\mathcal{T}_u$. Given these batches and the current threshold $\tau$, we instantiate Problem~\ref{eq:ot} with our choices of the domain-alignment cost and CRM-aware classification loss. For fixed model parameters $(g,h)$ and conformal threshold $\tau$, we optimize an OT coupling parameter $\gamma$ by solving

\begin{equation} \label{eq:ot}
\min_{\gamma \in \Pi(\mu_s, \mu_t)} \mathcal{L}_{\text{DA}}(\gamma, x^s, x^{u}; \theta) + \mathcal{L}_{\text{cls}}(\gamma,x^{u}, y^s;\theta,\tau)
\end{equation}

where 
\begin{align*}
    \mathcal{L}_{\text{DA}}(\gamma, x^s, x^{u}; \theta)
    &= \lambda_a \sum_{i,j=1}^{B} \gamma_{ij} 
    \underbrace{\left(c_{\mathrm{GOT}}(x_i^s,x_j^u)\right)}_{c\left(g(x_i^s),\, g(x_j^{u})\right)} 
    \\[6pt]
    \mathcal{L}_{\text{cls}}(\gamma,x^{u}, y^s;\theta,\tau)
    &= \lambda_{cls} \sum_{i,j=1}^{B} \gamma_{ij}
    \underbrace{\left(\sigma\left(\left(s\left(x_j^u, y_i^s\right) - \tau\right)   / T\right) \right)}_{\mathcal{L}_{cls}\left(y_i^s,\, h(g(x^{u}_j))\right)}.
\end{align*}

In Problem \ref{eq:ot}, $\sigma(\cdot)$ is the sigmoid function, and $s(\cdot, \cdot)$ is the nonconformity score of choice. In our implementation, we train using the HPS nonconformity score as $s(\cdot, \cdot)$, thus $s\left(x_j^u, y_i^s\right) := 1 - h_{y_i^s}\left(g(x_j^u)\right)$. The term $\sigma\left(s\left(x_j^u, y_i^s\right) - \tau \right)$ is a smoothed equivalent of the indicator function $\mathbbm{1}\left[s\left(x_j^u, y_i^s\right) > \tau\right]$. $T$ is a temperature hyperparameter that controls how closely the smooth approximation tracks the hard indicator; as $T \rightarrow 0$, it recovers the exact thresholding. Thus, $\mathcal{L}_{cls}$ defines a differentiable penalty when the conformal set of $x^{u}$ excludes the OT-propagated class $y_i^s$. 

The remaining term, $\mathcal{L}_{DA}$, is the domain-alignment loss in DeepJDOT's framework, which aligns the source and target domains. We replace the original squared Euclidean cost used in DeepJDOT with the label-guided Geometric Optimal Transport (GOT)~\citep{maman2025tmlr}, which incorporates source and target geometry through diffusion processes. We denote the resulting pairwise cost by $c_{\mathrm{GOT}}(x_i^s,x_j^u)$. The choice of transport cost is not central to our method; we use GOT as a state-of-the-art transport cost, while recognizing that the design of transport costs is itself an active area of research. We therefore omit the technical details here and briefly present GOT in Appendix~\ref{app:got} for completeness. The optimal coupling $\gamma$ obtained by solving the minimization Problem~\ref{eq:ot} serves a dual purpose: it aligns the source samples with the unlabeled target samples to enable the learning of a joint latent space, and it simultaneously assigns probabilistic pseudolabels to the unlabeled target samples. Given this fixed coupling $\gamma$, the algorithm then updates the model parameters $\theta$, comprising both the embedding function $g$ and the classifier $h$, via stochastic gradient descent to minimize the following empirical loss:

\begin{align} \label{eq:loss}
\mathcal{L}_{\text{total}}(\gamma, x^s,y^s,x^{u}; \theta, \tau) := \mathcal{L}_{\text{source}}(x^s,y^s;\theta) + \mathcal{L}_{\text{cls}}(\gamma,x^{u}, y^s;\theta,\tau)
&+ \mathcal{L}_{\text{DA}}(\gamma, x^s, x^{u}; \theta) 
+ \mathcal{L}_{\text{c}}(x^{u}; \theta, \tau)
\end{align}
where
\begin{align*} \label{eq:loss}
    &\mathcal{L}_{\text{source}}(x^s,y^s;\theta) = \frac{1}{B} \sum_{i=1}^{B} CE\left( y_i^s, h\left( g(x_i^s) \right) \right) \nonumber\\ 
    &\mathcal{L}_{\text{c}}(x^{u}; \theta, \tau) =\lambda_c \cdot \frac{1}{B}\sum_{j=1}^B \Omega\left(C(x^{u}_j; \theta, \tau)\right). \nonumber
\end{align*}

The $CE(\cdot)$ denotes the categorical cross-entropy function and $\Omega(C(x)) := \sum_{y \in \mathcal{Y}}\tilde{\mathbbm{1}}[s(x,y) \leq \tau]$ denotes the smoothed set size of a sample $x$, as also defined in Section \ref{section:prelim}. In \eqref{eq:loss}, $\mathcal{L}_{\text{source}}$ represents an Empirical Risk Minimization (ERM) objective on the labeled source domain, which is a standard objective in SSDA algorithms to avoid catastrophic forgetting of the source distribution. The term $\mathcal{L}_{\text{DA}}$ allows feature alignment between the source and target domains via Optimal Transport. To address the lack of target labels, $\mathcal{L}_{\text{cls}}$ acts as a cross-domain classification loss; it utilizes the coupling $\gamma$ to ensure that the label $y_i^s$ of the source sample $x_i^s$ that is matched to the target sample $x_j^{u}$ is properly included in the conformal set $C(x_j^{u})$. Finally, $\mathcal{L}_{\text{c}}$ is the penalty term used to minimize the average size of the prediction sets for the unlabeled target samples.  The full procedure is summarized in Algorithm~\ref{alg:ours}. In our implementation, we relax the marginal constraints in Problem~\ref{eq:ot} and compute the coupling using unbalanced Sinkhorn iterations. We provide the exact implementation details and solver settings in Appendix~\ref{app:implementation}.

\begin{algorithm}[t]
\caption{CP-JDOT}\label{alg:ours}
\begin{algorithmic}[1]
\State \textbf{Input:} $g, h, \mathcal{S}, \mathcal{T}_{\ell}, \mathcal{T}_u, 
\alpha, \lambda_\alpha, \lambda_c, \lambda_{cls}$
\State \textbf{Input:} Conformal Predictor (e.g. HPS) defining the nonconformity score $s(\cdot, \cdot)$.
\While{$g, h$ not converged}
    \State Calculate $\tau :=$ differentiable $(1-\alpha)$-quantile of the scores $s(x_i^l, y_i^l)$ using $\mathcal{T}_l$
    \State Randomly sample $\mathcal{B}_s \subset \mathcal{S}$ and $\mathcal{B}_t \subset \mathcal{T}_u$
    \State Freeze $g, h$ and solve Problem \ref{eq:ot}  for $\gamma$, using $\tau, \mathcal{B}_s, \mathcal{B}_t$.
    \State Compute the loss in \eqref{eq:loss} using $\gamma$ and the conformal sets. \State Update $g, h$ via backpropagation.
\EndWhile
\State \textbf{Output:} Trained $g, h$.
\end{algorithmic}
\end{algorithm}

\subsection{Encoding Domain Knowledge into
    \texorpdfstring{$\mathcal{L}_{\text{cls}}$}{L\_cls}}
\label{section:control}

The loss function $\mathcal{L}_{\text{cls}}$ in \eqref{eq:loss} encourages that the conformal set includes the correct class, which represents the primary requirement for maintaining set validity. In practice, however, it is often desirable to exercise finer control over the composition of the conformal sets, even if this results in a slight reduction in set efficiency (cardinality). For instance, in a skin lesion classification task, a conformal set containing only benign classes --- such as \{melanocytic nevi, dermatofibroma\} --- may be preferable to an equal-size set that mixes diagnostic categories, such as \{melanoma, melanocytic nevi\} (one malignant and one benign), since the latter simultaneously suggests a cancer-positive and a cancer-negative diagnosis. Similar to the Conformal Training approach by \cite{stutz2021learning}, to accommodate such hierarchical constraints where classes belong to distinct parent categories, we replace $\mathcal{L}_{\text{cls}}$ in \eqref{eq:loss} with a specialized control loss, $\mathcal{L}_{\text{control}}$, defined as:

\begin{equation} \label{eq:special-loss}
\mathcal{L}_{\text{control}}(\gamma,x^{u}, y^s;\theta,\tau) := \lambda_{ctrl} \sum_{i,j=1}^{B} \gamma_{ij}\Big(\tilde{\mathbbm{1}}\{ y_i^s \notin C(x_j^u; \theta, \tau) \} 
+ \lambda \sum_{y \in \mathcal{Y}_{\text{err}}} \tilde{\mathbbm{1}}\{ y \in C(x_j^u; \theta, \tau) \}\Big),
\end{equation}

where $\mathcal{Y}_{\text{err}}$ represents the set of conflicting classes given the transported label $y_i^s$. The first term ensures the transported label is included in the set, while the second term penalizes the inclusion of ``conflicting'' classes. Thus, our final objective when we are given such structural constraints is:

\begin{align*} \label{eq:special-total}
    \mathcal{L}_{\text{total,ctrl}}(\gamma, x^s,y^s,x^{u}; \theta, \tau) := \mathcal{L}_{\text{source}}(x^s,y^s;\theta) + \mathcal{L}_{\text{DA}}(\gamma, x^s, x^{u}; \theta) &+ \mathcal{L}_{\text{control}}(\gamma,x^{u}, y^s;\theta,\tau) \\
&+ \mathcal{L}_{\text{c}}(x^{u}; \theta, \tau).
\end{align*}

%% file: sections/results.tex
Our experiments are organized as follows. In Section~\ref{exp:size}, we compare the conformal set sizes produced by our method with those of state-of-the-art SSDA methods designed for top-1 accuracy, calibrated using post-hoc split CP with the HPS and APS scores presented in Section~\ref{section:prelim}. In Section~\ref{exp:ablation}, we perform an ablation study to confirm that the set size gains stem from the proposed conformal objective. In Section~\ref{exp:control}, we show how $\mathcal{L}_{\text{control}}$, the alternative loss described in Section~\ref{section:control}, can influence the structure of prediction sets when domain-specific structural constraints are available. In Section~\ref{exp:alpha}, we keep training-time $\alpha$ fixed and vary test-time $\alpha$ to assess whether a single trained model retains its set size gains across coverage levels. This is important because our method requires choosing $\alpha$ during training, when we may not know what coverage level will be required at deployment. Finally, in Section~\ref{exp:data}, we examine how our algorithm performs under different numbers of labeled target examples and choices of $\alpha$. Small values of $\alpha$ with few labeled examples can make the training-time quantile estimate depend on just a few extreme scores, potentially making it unstable. We perform a sweep over both quantities to characterize where our method yields smaller sets and where the baseline algorithms may be preferable. Throughout the experimental section, in every table, the best entry is in bold, and the second-best entry is underlined.

We use three benchmark Domain Adaptation datasets in our experiments. \textbf{Office-Home} \citep{venkateswara2017deep} is a standard domain adaptation benchmark consisting of images from four domains: Art (A), Clipart (C), Product (P), and Real World (R), with 65 object categories. We evaluate all 12 pairwise transfer directions using a ResNet-34 backbone \citep{HeCVPR2016} in the 1-shot and 3-shot settings, where only one and three labeled target samples per class are available, respectively. \textbf{DomainNet126} follows the SSDA setup of \citet{saito2019semi}, using 126 object categories from four domains: Real (R), Clipart (C), Painting (P), and Sketch (S). We evaluate seven transfer directions using a ResNet-50 backbone in the 1-shot and 3-shot settings. \textbf{Skin Lesion} defines a challenging medical imaging transfer learning setting, where HAM10000 \citep{tschandl2018ham10000} always serves as the labeled source domain and five target domains of varying difficulty are considered: PH2 \citep{ph2}, Derm7pt \citep{derm7pt}, and 3 datasets from the ISIC repository \citep{isic} --- MSK, SONIC and UDA. In each experiment, one of PH2, MSK, UDA, SONIC, or Derm7pt is the target domain, and we pool HAM10000 with the remaining 4 datasets from this group to form the labeled source dataset. We use a ResNet-18 backbone for this setting with 10 labeled target samples per class. For all experiments, results are averaged over 10 runs with 10 random calibration-test splits, using the same trained model, to account for variance stemming from the post-hoc calibration sample. For each run, the target dataset's test split is further split 50/50: the first half is used as the held-out conformal calibration set, and the remaining target samples form the test set.

\begin{table*}[t!]
\centering
\resizebox{\textwidth}{!}{
\begin{tabular}{lccccccccccccc}
\toprule
Method
& A$\rightarrow$C & A$\rightarrow$P & A$\rightarrow$R
& C$\rightarrow$A & C$\rightarrow$P & C$\rightarrow$R
& P$\rightarrow$A & P$\rightarrow$C & P$\rightarrow$R
& R$\rightarrow$A & R$\rightarrow$C & R$\rightarrow$P
& Avg \\
\midrule
S\&T & $21.52 \pm 0.99$ & $6.93 \pm 0.41$ & $5.08 \pm 0.37$ & $13.45 \pm 0.91$ & $8.25 \pm 0.72$ & $6.67 \pm 0.39$ & $17.30 \pm 1.84$ & $26.54 \pm 1.85$ & $4.11 \pm 0.38$ & $9.81 \pm 1.21$ & $20.11 \pm 1.55$ & $4.93 \pm 0.43$ & $12.06$ \\
\midrule
MME & $20.49 \pm 0.89$ & $6.64 \pm 0.46$ & $4.72 \pm 0.31$ & $12.20 \pm 0.82$ & $7.82 \pm 0.68$ & $6.25 \pm 0.37$ & $16.79 \pm 1.72$ & $25.64 \pm 1.78$ & $3.85 \pm 0.33$ & $9.23 \pm 1.15$ & $19.15 \pm 1.42$ & $4.56 \pm 0.39$ & $11.45$ \\
CDAC & $20.04 \pm 0.88$ & $6.37 \pm 0.39$ & $4.70 \pm 0.35$ & $11.82 \pm 0.79$ & $7.48 \pm 0.61$ & $6.07 \pm 0.34$ & $16.34 \pm 1.69$ & $25.20 \pm 1.71$ & $3.77 \pm 0.30$ & $9.02 \pm 1.06$ & $18.99 \pm 1.44$ & $4.42 \pm 0.37$ & $11.19$ \\
DECOTA & $21.04 \pm 0.83$ & $7.39 \pm 0.35$ & $5.67 \pm 0.28$ & $12.66 \pm 0.75$ & $8.53 \pm 0.60$ & $7.11 \pm 0.35$ & $17.10 \pm 1.64$ & $26.22 \pm 1.66$ & $4.91 \pm 0.29$ & $10.28 \pm 1.02$ & $19.89 \pm 1.36$ & $5.48 \pm 0.35$ & $12.19$ \\
IDMNE & $19.15 \pm 0.78$ & $6.26 \pm 0.37$ & $4.36 \pm 0.27$ & $10.96 \pm 0.69$ & $7.07 \pm 0.54$ & $5.59 \pm 0.29$ & $\mathbf{14.48 \pm 1.61}$ & $24.08 \pm 1.55$ & $3.61 \pm 0.25$ & $8.34 \pm 0.95$ & $17.95 \pm 1.30$ & $4.14 \pm 0.34$ & $10.50$ \\
ProML & $19.56 \pm 0.81$ & $6.00 \pm 0.32$ & $4.22 \pm 0.29$ & $11.22 \pm 0.72$ & $6.75 \pm 0.48$ & $5.82 \pm 0.32$ & $15.82 \pm 1.74$ & $24.47 \pm 1.63$ & $3.67 \pm 0.28$ & $8.55 \pm 0.96$ & $18.36 \pm 1.27$ & $3.91 \pm 0.31$ & $10.70$ \\
EFTL & \underline{$19.04 \pm 0.75$} & \underline{$5.93 \pm 0.30$} & \underline{$4.15 \pm 0.26$} & \underline{$10.72 \pm 0.66$} & \underline{$6.68 \pm 0.43$} & \underline{$5.53 \pm 0.28$} & $14.76 \pm 1.62$ & $\mathbf{23.74 \pm 1.52}$ & \underline{$3.52 \pm 0.24$} & \underline{$8.12 \pm 0.92$} & \underline{$17.83 \pm 1.22$} & \underline{$3.75 \pm 0.28$} & \underline{$10.31$} \\
\midrule
CP-JDOT & $\mathbf{18.68 \pm 0.65}$ & $\mathbf{5.59 \pm 0.24}$ & $\mathbf{3.84 \pm 0.21}$ & $\mathbf{10.06 \pm 0.51}$ & $\mathbf{6.15 \pm 0.36}$ & $\mathbf{5.21 \pm 0.23}$ & \underline{$14.53 \pm 1.56$} & \underline{$23.87 \pm 1.46$} & $\mathbf{3.22 \pm 0.21}$ & $\mathbf{7.67 \pm 0.83}$ & $\mathbf{17.72 \pm 1.13}$ & $\mathbf{3.33 \pm 0.23}$ & $\mathbf{9.99}$ \\
\bottomrule
\end{tabular}
}
\caption{Avg. Set Size on Office-Home (ResNet-34 / 1-Shot / $\alpha = 0.05$ / HPS)}
\label{tab:office_home_1shot_005_hps}
\end{table*}

\begin{table*}[t!]
\centering
\resizebox{\textwidth}{!}{
\begin{tabular}{lccccccccccccc}
\toprule
Method
& A$\rightarrow$C & A$\rightarrow$P & A$\rightarrow$R
& C$\rightarrow$A & C$\rightarrow$P & C$\rightarrow$R
& P$\rightarrow$A & P$\rightarrow$C & P$\rightarrow$R
& R$\rightarrow$A & R$\rightarrow$C & R$\rightarrow$P
& Avg \\
\midrule
S\&T & $34.20 \pm 1.35$ & $19.12 \pm 1.07$ & $14.64 \pm 0.83$ & $24.93 \pm 1.12$ & $22.13 \pm 1.20$ & $16.10 \pm 0.71$ & $27.00 \pm 1.59$ & $37.32 \pm 2.14$ & $13.84 \pm 0.71$ & $21.80 \pm 1.08$ & $30.26 \pm 1.47$ & $15.53 \pm 0.69$ & $23.07$ \\
\midrule
MME & $30.36 \pm 1.22$ & $16.94 \pm 0.98$ & $13.03 \pm 0.76$ & $21.98 \pm 0.99$ & $18.87 \pm 1.08$ & $14.36 \pm 0.60$ & $24.99 \pm 1.42$ & $32.60 \pm 2.03$ & $12.35 \pm 0.66$ & $19.70 \pm 0.96$ & $27.49 \pm 1.36$ & $12.70 \pm 0.62$ & $20.45$ \\
CDAC & $29.11 \pm 1.14$ & $15.59 \pm 0.94$ & $11.59 \pm 0.72$ & $20.19 \pm 0.94$ & $17.68 \pm 0.96$ & $13.40 \pm 0.56$ & $23.72 \pm 1.44$ & $31.48 \pm 1.96$ & $11.77 \pm 0.60$ & $17.89 \pm 0.91$ & $26.24 \pm 1.21$ & $11.66 \pm 0.56$ & $19.19$ \\
DECOTA & $32.95 \pm 1.20$ & $19.25 \pm 0.90$ & $15.38 \pm 0.70$ & $22.83 \pm 0.90$ & $20.07 \pm 0.93$ & $16.84 \pm 0.62$ & $27.02 \pm 1.36$ & $35.16 \pm 1.92$ & $15.43 \pm 0.62$ & $21.56 \pm 0.84$ & $28.30 \pm 1.24$ & $15.26 \pm 0.60$ & $22.50$ \\
IDMNE & $26.18 \pm 1.02$ & $13.41 \pm 0.85$ & $10.39 \pm 0.66$ & $16.85 \pm 0.81$ & $15.01 \pm 0.89$ & $11.05 \pm 0.52$ & $20.41 \pm 1.30$ & $27.97 \pm 1.81$ & $9.55 \pm 0.53$ & $15.16 \pm 0.79$ & $21.85 \pm 1.05$ & $10.16 \pm 0.49$ & $16.50$ \\
ProML & $27.79 \pm 1.06$ & \underline{$12.84 \pm 0.81$} & $9.76 \pm 0.62$ & $18.11 \pm 0.83$ & $13.82 \pm 0.85$ & $12.50 \pm 0.58$ & $22.18 \pm 1.35$ & $29.18 \pm 1.87$ & $10.02 \pm 0.56$ & $16.11 \pm 0.85$ & $23.52 \pm 1.13$ & $9.22 \pm 0.45$ & $17.09$ \\
EFTL & \underline{$24.99 \pm 0.96$} & $13.03 \pm 0.78$ & \underline{$9.42 \pm 0.63$} & \underline{$16.21 \pm 0.76$} & \underline{$12.97 \pm 0.79$} & \underline{$10.80 \pm 0.47$} & \underline{$19.70 \pm 1.27$} & $\mathbf{26.34 \pm 1.78}$ & \underline{$9.00 \pm 0.48$} & \underline{$14.22 \pm 0.72$} & \underline{$21.27 \pm 1.02$} & \underline{$8.88 \pm 0.42$} & \underline{$15.57$} \\
\midrule
CP-JDOT & $\mathbf{23.28 \pm 0.83}$ & $\mathbf{11.13 \pm 0.70}$ & $\mathbf{8.22 \pm 0.57}$ & $\mathbf{14.18 \pm 0.63}$ & $\mathbf{11.51 \pm 0.66}$ & $\mathbf{9.71 \pm 0.38}$ & $\mathbf{18.60 \pm 1.14}$ & \underline{$26.44 \pm 1.70$} & $\mathbf{7.65 \pm 0.43}$ & $\mathbf{12.64 \pm 0.64}$ & $\mathbf{20.59 \pm 0.90}$ & $\mathbf{7.22 \pm 0.36}$ & $\mathbf{14.26}$ \\
\bottomrule
\end{tabular}
}
\caption{Avg. Set Size on Office-Home (ResNet-34 / 1-Shot / $\alpha = 0.05$ / APS)}
\label{tab:office_home_1shot_005_aps}
\end{table*}

\subsection{Average Set Size Reduction} \label{exp:size}

In this section, we evaluate whether CP-JDOT can reduce conformal set size relative to state-of-the-art SSDA methods followed by post-hoc conformal calibration. We compare against the following baselines: Empirical Risk Minimization on the labeled source and target samples (S\&T), MME~\citep{saito2019semi}, CDAC~\citep{li2021cdac}, DECOTA~\citep{yang2021decota}, ProML~\citep{huang2023promm}, EFTL~\citep{he2024eftl}, and IDMNE~\citep{li2024idmne}. These methods span a wide range of SSDA approaches, including adversarial learning (MME and CDAC), co-training with task decomposition (DECOTA), and prototype- and metric-learning-based approaches (ProML and IDMNE). To provide a consistent comparison, we conformalize each baseline using the Homogeneous Prediction Sets~\citep{SadinleJASA2019} and the Adaptive Prediction Set (APS) scores~\citep{RomanoNIPS2020} post-hoc. Using HPS as the nonconformity score leads to the smallest possible sets that satisfy the coverage constraint in expectation, but has the downside of sometimes producing empty sets. APS leads to more inflated prediction sets but empirically achieves better conditional coverage, thus the two scores are complementary and provide a well-rounded evaluation. Since coverage is guaranteed because of the test-time recalibration, we generally omit the coverage tables for brevity. We provide 3 coverage tables for the HPS score and $\alpha \in \{0.025, 0.05, 0.1\}$ in Appendix \ref{additional:coverage}. 

Tables~\ref{tab:office_home_1shot_005_hps}, \ref{tab:office_home_1shot_005_aps}, \ref{tab:office_home_3shot_005_hps} and \ref{tab:office_home_3shot_005_aps} report average set sizes across all 12 Office-Home transfer directions under the 1-shot and 3-shot settings with a ResNet-34 backbone, evaluated with the HPS and APS nonconformity scores at miscoverage level $\alpha=0.05$. CP-JDOT consistently achieves the smallest average set size across 43 of 48 transfer directions, outperforming the best-performing baseline by 3.1\% and 4.5\% for HPS, 8.4\% and 8.1\% for APS, in 1-shot and 3-shot respectively. We observe that the improvement is most pronounced in easier transfer directions such as A$\rightarrow$R and P$\rightarrow$R. These directions generally exhibit smaller domain shifts and higher baseline performance, which yields more reliable OT-derived pseudolabels and allows the conformal alignment objective to more effectively exploit the available target data. In a few harder transfer directions such as A$\rightarrow$C and P$\rightarrow$A, where most methods achieve lower accuracy, the gains are smaller but consistent, suggesting that conformal-aware training provides a benefit even when the OT-derived pseudolabels are less accurate. One major exception is $P\rightarrow C$, the hardest task, where Optimal Transport induces severe enough negative transfer that our conformal objective cannot beat the best baseline. We provide tables with the top-1 accuracy of each algorithm in Appendix \ref{additional:acc}. Notably, CP-JDOT improves conformal set size despite achieving up to 8.7\% \emph{lower} top-1 accuracy than the strongest SSDA baselines in some transfer directions. This finding is consistent with observations from the supervised CRM literature~\citep{stutz2021learning,einbinder2022cut,shi2025dpsm} and supports the hypothesis that the conventional top-1 accuracy objective can be misaligned with the downstream conformal set-size efficiency objective.

\begin{table*}[t!]
\centering
\resizebox{\textwidth}{!}{
\begin{tabular}{lccccccccccccc}
\toprule
Method
& A$\rightarrow$C & A$\rightarrow$P & A$\rightarrow$R
& C$\rightarrow$A & C$\rightarrow$P & C$\rightarrow$R
& P$\rightarrow$A & P$\rightarrow$C & P$\rightarrow$R
& R$\rightarrow$A & R$\rightarrow$C & R$\rightarrow$P
& Avg \\
\midrule
S\&T & $16.89 \pm 0.97$ & $5.12 \pm 0.29$ & $4.02 \pm 0.33$ & $11.24 \pm 0.95$ & $6.07 \pm 0.53$ & $4.98 \pm 0.36$ & $14.63 \pm 1.12$ & $18.66 \pm 1.19$ & $3.72 \pm 0.34$ & $8.34 \pm 0.84$ & $14.89 \pm 0.76$ & $3.08 \pm 0.23$ & $9.30$ \\
\midrule
MME & $16.13 \pm 0.91$ & $4.89 \pm 0.25$ & $3.76 \pm 0.29$ & $10.58 \pm 0.90$ & $5.73 \pm 0.47$ & $4.69 \pm 0.32$ & $13.92 \pm 1.02$ & $17.96 \pm 1.08$ & $3.49 \pm 0.30$ & $7.84 \pm 0.76$ & $14.33 \pm 0.69$ & $2.84 \pm 0.20$ & $8.85$ \\
CDAC & $15.88 \pm 0.86$ & $4.65 \pm 0.22$ & $3.67 \pm 0.26$ & $10.41 \pm 0.85$ & $5.54 \pm 0.43$ & $4.53 \pm 0.28$ & $13.56 \pm 0.97$ & $17.68 \pm 0.99$ & $3.38 \pm 0.29$ & $7.39 \pm 0.69$ & $14.07 \pm 0.61$ & $2.70 \pm 0.17$ & $8.62$ \\
DECOTA & $15.72 \pm 0.90$ & $4.74 \pm 0.20$ & $3.52 \pm 0.28$ & $10.23 \pm 0.87$ & $5.40 \pm 0.46$ & $4.59 \pm 0.30$ & $13.21 \pm 0.92$ & $17.16 \pm 0.96$ & $3.41 \pm 0.27$ & $7.54 \pm 0.72$ & $13.79 \pm 0.64$ & $2.66 \pm 0.18$ & $8.50$ \\
IDMNE & \underline{$15.17 \pm 0.81$} & $4.53 \pm 0.18$ & $3.46 \pm 0.23$ & $9.92 \pm 0.81$ & $5.23 \pm 0.39$ & \underline{$4.24 \pm 0.25$} & \underline{$12.28 \pm 0.82$} & $16.63 \pm 0.91$ & \underline{$3.18 \pm 0.26$} & $7.04 \pm 0.66$ & $13.40 \pm 0.56$ & $2.53 \pm 0.14$ & $8.13$ \\
ProML & $15.46 \pm 0.85$ & \underline{$4.38 \pm 0.16$} & \underline{$3.29 \pm 0.24$} & $10.06 \pm 0.88$ & \underline{$5.08 \pm 0.38$} & $4.39 \pm 0.27$ & $12.78 \pm 0.86$ & $16.97 \pm 0.94$ & $3.24 \pm 0.25$ & $7.12 \pm 0.63$ & $13.52 \pm 0.58$ & \underline{$2.44 \pm 0.13$} & $8.23$ \\
EFTL & $15.28 \pm 0.77$ & $4.46 \pm 0.15$ & $3.38 \pm 0.22$ & \underline{$9.86 \pm 0.79$} & $5.16 \pm 0.36$ & $4.31 \pm 0.23$ & $12.59 \pm 0.79$ & $\mathbf{15.63 \pm 0.83}$ & $3.21 \pm 0.23$ & \underline{$7.02 \pm 0.61$} & \underline{$13.32 \pm 0.53$} & $2.48 \pm 0.12$ & \underline{$8.06$} \\
\midrule
CP-JDOT & $\mathbf{14.60 \pm 0.80}$ & $\mathbf{4.14 \pm 0.13}$ & $\mathbf{3.17 \pm 0.22}$ & $\mathbf{9.40 \pm 0.80}$ & $\mathbf{4.85 \pm 0.37}$ & $\mathbf{4.01 \pm 0.22}$ & $\mathbf{11.90 \pm 0.76}$ & \underline{$15.75 \pm 0.78$} & $\mathbf{2.99 \pm 0.24}$ & $\mathbf{6.49 \pm 0.60}$ & $\mathbf{12.79 \pm 0.48}$ & $\mathbf{2.33 \pm 0.10}$ & $\mathbf{7.70}$ \\
\bottomrule
\end{tabular}
}
\caption{Avg. Set Size on Office-Home (ResNet-34 / 3-Shot / $\alpha = 0.05$ / HPS)}
\label{tab:office_home_3shot_005_hps}
\end{table*}

\begin{table*}[t!]
\centering
\resizebox{\textwidth}{!}{
\begin{tabular}{lccccccccccccc}
\toprule
Method
& A$\rightarrow$C & A$\rightarrow$P & A$\rightarrow$R
& C$\rightarrow$A & C$\rightarrow$P & C$\rightarrow$R
& P$\rightarrow$A & P$\rightarrow$C & P$\rightarrow$R
& R$\rightarrow$A & R$\rightarrow$C & R$\rightarrow$P
& Avg \\
\midrule
S\&T & $24.57 \pm 1.30$ & $12.05 \pm 0.78$ & $9.74 \pm 0.63$ & $18.11 \pm 1.47$ & $13.32 \pm 0.83$ & $11.10 \pm 0.54$ & $21.60 \pm 1.67$ & $24.10 \pm 1.25$ & $9.73 \pm 0.55$ & $15.99 \pm 1.08$ & $21.72 \pm 0.89$ & $8.64 \pm 0.60$ & $15.89$ \\
\midrule
MME & $23.01 \pm 1.22$ & $11.24 \pm 0.70$ & $9.13 \pm 0.57$ & $17.50 \pm 1.32$ & $12.67 \pm 0.74$ & $10.24 \pm 0.48$ & $20.75 \pm 1.59$ & $23.16 \pm 1.17$ & $8.94 \pm 0.51$ & $15.14 \pm 0.96$ & $20.68 \pm 0.82$ & $7.93 \pm 0.53$ & $15.03$ \\
CDAC & $22.58 \pm 1.16$ & $10.86 \pm 0.63$ & $8.48 \pm 0.51$ & $16.87 \pm 1.26$ & $12.14 \pm 0.66$ & $9.96 \pm 0.43$ & $20.08 \pm 1.51$ & $22.54 \pm 1.04$ & $8.43 \pm 0.44$ & $14.78 \pm 0.91$ & $19.97 \pm 0.73$ & $7.60 \pm 0.48$ & $14.52$ \\
DECOTA & $22.79 \pm 1.18$ & $10.47 \pm 0.57$ & $8.71 \pm 0.53$ & $16.49 \pm 1.28$ & $11.87 \pm 0.62$ & $10.07 \pm 0.46$ & $19.64 \pm 1.48$ & $22.19 \pm 0.98$ & $8.61 \pm 0.47$ & $14.51 \pm 0.87$ & $19.53 \pm 0.67$ & $7.46 \pm 0.46$ & $14.36$ \\
IDMNE & $21.48 \pm 1.06$ & $10.11 \pm 0.52$ & $8.03 \pm 0.47$ & $15.66 \pm 1.18$ & $11.51 \pm 0.58$ & $9.41 \pm 0.37$ & $18.84 \pm 1.39$ & $21.69 \pm 0.88$ & $7.96 \pm 0.40$ & $13.89 \pm 0.83$ & $18.63 \pm 0.58$ & $7.04 \pm 0.43$ & $13.69$ \\
ProML & $21.91 \pm 1.10$ & \underline{$9.73 \pm 0.48$} & \underline{$7.85 \pm 0.42$} & $15.92 \pm 1.24$ & \underline{$11.09 \pm 0.54$} & $9.66 \pm 0.41$ & $19.38 \pm 1.45$ & $21.28 \pm 0.92$ & $8.18 \pm 0.43$ & $14.04 \pm 0.86$ & $18.95 \pm 0.63$ & \underline{$6.74 \pm 0.39$} & $13.73$ \\
EFTL & \underline{$21.29 \pm 1.01$} & $10.00 \pm 0.46$ & $7.94 \pm 0.43$ & \underline{$15.32 \pm 1.14$} & $11.27 \pm 0.51$ & \underline{$9.38 \pm 0.35$} & \underline{$18.75 \pm 1.36$} & $\mathbf{19.71 \pm 0.75}$ & \underline{$7.88 \pm 0.36$} & \underline{$13.76 \pm 0.80$} & \underline{$18.49 \pm 0.54$} & $6.76 \pm 0.38$ & \underline{$13.38$} \\
\midrule
CP-JDOT & $\mathbf{19.36 \pm 0.95}$ & $\mathbf{8.76 \pm 0.41}$ & $\mathbf{7.11 \pm 0.40}$ & $\mathbf{14.23 \pm 1.20}$ & $\mathbf{9.92 \pm 0.48}$ & $\mathbf{8.53 \pm 0.30}$ & $\mathbf{17.39 \pm 1.34}$ & \underline{$19.89 \pm 0.68$} & $\mathbf{6.93 \pm 0.34}$ & $\mathbf{12.23 \pm 0.78}$ & $\mathbf{17.13 \pm 0.47}$ & $\mathbf{6.07 \pm 0.37}$ & $\mathbf{12.30}$ \\
\bottomrule
\end{tabular}
}
\caption{Avg. Set Size on Office-Home (ResNet-34 / 3-Shot / $\alpha = 0.05$ / APS)}
\label{tab:office_home_3shot_005_aps}
\end{table*}

Tables~\ref{tab:domainnet126_1shot_005_hps}, \ref{tab:domainnet126_1shot_005_aps}, \ref{tab:domainnet126_3shot_005_hps} and \ref{tab:domainnet126_3shot_005_aps} report average set sizes across the seven DomainNet126 transfer directions under the 1-shot and 3-shot settings with a ResNet-50 backbone, evaluated with the HPS and APS nonconformity scores at miscoverage level $\alpha=0.05$. CP-JDOT achieves the smallest average set size in 26 of the 28 combinations of transfer direction, shot count, and score. It outperforms the best-performing baseline on average, IDMNE, by 4.0\% and 5.2\% for HPS and by 9.2\% and 8.9\% for APS, in the 1-shot and 3-shot settings, respectively. As in Office-Home, we observe gains in easier transfer directions such as P$\rightarrow$R, where most methods achieve higher top-1 accuracy. Similarly as in Office-Home, the only task where CP-JDOT loses to a baseline is the harder C$\rightarrow$S task in the 1-shot setting, where IDMNE produces slightly smaller sets for both HPS and APS.

\begin{table*}[t!]
\centering
\resizebox{\textwidth}{!}{
\begin{tabular}{lcccccccc}
\toprule
Method
& R$\rightarrow$C & R$\rightarrow$P & P$\rightarrow$C
& C$\rightarrow$S & S$\rightarrow$P & R$\rightarrow$S & P$\rightarrow$R
& Avg \\
\midrule
S\&T & $9.30 \pm 0.71$ & $11.91 \pm 0.90$ & $9.84 \pm 0.75$ & $25.57 \pm 1.89$ & $15.20 \pm 1.14$ & $18.75 \pm 1.40$ & $6.10 \pm 0.47$ & $13.81$ \\
\midrule
MME & $8.70 \pm 0.66$ & $11.49 \pm 0.86$ & $9.50 \pm 0.71$ & $24.50 \pm 1.79$ & $14.54 \pm 1.08$ & $18.20 \pm 1.34$ & $5.97 \pm 0.46$ & $13.27$ \\
CDAC & $8.01 \pm 0.59$ & $10.63 \pm 0.78$ & $8.80 \pm 0.65$ & $21.60 \pm 1.55$ & $13.46 \pm 0.98$ & $16.54 \pm 1.19$ & $5.66 \pm 0.42$ & $12.10$ \\
DECOTA & \underline{$7.33 \pm 0.53$} & \underline{$9.78 \pm 0.69$} & $8.07 \pm 0.58$ & $20.37 \pm 1.41$ & $12.33 \pm 0.87$ & $15.19 \pm 1.06$ & $5.31 \pm 0.38$ & $11.20$ \\
IDMNE & $7.48 \pm 0.52$ & $9.94 \pm 0.69$ & $8.10 \pm 0.56$ & $\mathbf{19.69 \pm 1.33}$ & \underline{$12.28 \pm 0.84$} & \underline{$15.07 \pm 1.03$} & $5.33 \pm 0.38$ & \underline{$11.13$} \\
ProML & $7.70 \pm 0.54$ & $10.20 \pm 0.71$ & $8.37 \pm 0.58$ & $20.41 \pm 1.38$ & $12.69 \pm 0.87$ & $15.55 \pm 1.06$ & $5.35 \pm 0.38$ & $11.47$ \\
EFTL & $7.35 \pm 0.49$ & $9.85 \pm 0.65$ & \underline{$8.04 \pm 0.53$} & \underline{$19.75 \pm 1.27$} & $12.44 \pm 0.81$ & $15.27 \pm 0.99$ & \underline{$5.20 \pm 0.35$} & \underline{$11.13$} \\
\midrule
CP-JDOT & $\mathbf{6.76 \pm 0.41}$ & $\mathbf{9.22 \pm 0.55}$ & $\mathbf{7.45 \pm 0.45}$ & $19.78 \pm 1.15$ & $\mathbf{12.03 \pm 0.71}$ & $\mathbf{14.94 \pm 0.88}$ & $\mathbf{4.62 \pm 0.28}$ & $\mathbf{10.69}$ \\
\bottomrule
\end{tabular}
}
\caption{Avg. Set Size on DomainNet126 (ResNet-50 / 1-Shot / $\alpha = 0.05$ / HPS)}
\label{tab:domainnet126_1shot_005_hps}
\end{table*}

\begin{table*}[t!]
\centering
\resizebox{\textwidth}{!}{
\begin{tabular}{lcccccccc}
\toprule
Method
& R$\rightarrow$C & R$\rightarrow$P & P$\rightarrow$C
& C$\rightarrow$S & S$\rightarrow$P & R$\rightarrow$S & P$\rightarrow$R
& Avg \\
\midrule
S\&T & $32.43 \pm 1.59$ & $34.37 \pm 1.68$ & $31.41 \pm 1.54$ & $55.92 \pm 2.65$ & $40.16 \pm 1.94$ & $44.85 \pm 2.15$ & $22.54 \pm 1.13$ & $37.38$ \\
\midrule
MME & $28.08 \pm 1.41$ & $32.48 \pm 1.61$ & $29.69 \pm 1.48$ & $52.07 \pm 2.52$ & $37.15 \pm 1.83$ & $43.16 \pm 2.11$ & $22.00 \pm 1.12$ & $34.95$ \\
CDAC & $22.40 \pm 1.14$ & $26.41 \pm 1.33$ & $24.16 \pm 1.22$ & $37.77 \pm 1.86$ & $30.39 \pm 1.51$ & $33.68 \pm 1.67$ & $18.87 \pm 0.97$ & $27.67$ \\
DECOTA & $19.11 \pm 0.96$ & $23.02 \pm 1.14$ & $20.74 \pm 1.03$ & $35.39 \pm 1.71$ & $26.30 \pm 1.29$ & $29.46 \pm 1.44$ & $17.05 \pm 0.86$ & $24.44$ \\
IDMNE & $18.87 \pm 1.00$ & $22.43 \pm 1.18$ & \underline{$19.58 \pm 1.04$} & $\mathbf{30.36 \pm 1.57}$ & \underline{$24.30 \pm 1.27$} & \underline{$26.97 \pm 1.40$} & $16.10 \pm 0.86$ & \underline{$22.66$} \\
ProML & $20.66 \pm 1.09$ & $24.29 \pm 1.26$ & $21.67 \pm 1.14$ & $33.59 \pm 1.72$ & $26.74 \pm 1.39$ & $29.56 \pm 1.52$ & $16.46 \pm 0.88$ & $24.71$ \\
EFTL & \underline{$18.38 \pm 0.96$} & \underline{$22.39 \pm 1.16$} & $19.59 \pm 1.02$ & $31.23 \pm 1.59$ & $25.69 \pm 1.32$ & $28.52 \pm 1.46$ & \underline{$15.34 \pm 0.81$} & $23.02$ \\
\midrule
CP-JDOT & $\mathbf{15.92 \pm 0.82}$ & $\mathbf{19.98 \pm 1.01}$ & $\mathbf{17.17 \pm 0.88}$ & \underline{$30.37 \pm 1.50$} & $\mathbf{22.28 \pm 1.12}$ & $\mathbf{25.88 \pm 1.29}$ & $\mathbf{12.45 \pm 0.65}$ & $\mathbf{20.58}$ \\
\bottomrule
\end{tabular}
}
\caption{Avg. Set Size on DomainNet126 (ResNet-50 / 1-Shot / $\alpha = 0.05$ / APS)}
\label{tab:domainnet126_1shot_005_aps}
\end{table*}

\begin{table*}[t!]
\centering
\resizebox{\textwidth}{!}{
\begin{tabular}{lcccccccc}
\toprule
Method
& R$\rightarrow$C & R$\rightarrow$P & P$\rightarrow$C
& C$\rightarrow$S & S$\rightarrow$P & R$\rightarrow$S & P$\rightarrow$R
& Avg \\
\midrule
S\&T & $7.56 \pm 0.56$ & $10.59 \pm 0.76$ & $7.74 \pm 0.57$ & $23.83 \pm 1.63$ & $13.93 \pm 0.98$ & $18.05 \pm 1.25$ & $4.66 \pm 0.35$ & $12.34$ \\
\midrule
MME & $7.10 \pm 0.50$ & $10.35 \pm 0.71$ & $7.55 \pm 0.53$ & $22.26 \pm 1.44$ & $13.50 \pm 0.91$ & $18.06 \pm 1.19$ & $4.63 \pm 0.33$ & $11.92$ \\
CDAC & $6.43 \pm 0.42$ & $9.55 \pm 0.61$ & $6.81 \pm 0.45$ & $19.74 \pm 1.21$ & $12.25 \pm 0.77$ & $15.54 \pm 0.97$ & $4.41 \pm 0.30$ & $10.68$ \\
DECOTA & $6.37 \pm 0.43$ & $9.54 \pm 0.63$ & $6.86 \pm 0.46$ & $20.08 \pm 1.26$ & $12.36 \pm 0.80$ & $15.66 \pm 1.00$ & $4.43 \pm 0.31$ & $10.76$ \\
IDMNE & \underline{$6.07 \pm 0.38$} & \underline{$8.91 \pm 0.54$} & \underline{$6.44 \pm 0.40$} & \underline{$18.23 \pm 1.06$} & \underline{$11.38 \pm 0.68$} & \underline{$14.53 \pm 0.86$} & $4.19 \pm 0.27$ & \underline{$9.96$} \\
ProML & $6.22 \pm 0.39$ & $9.12 \pm 0.56$ & $6.66 \pm 0.42$ & $18.64 \pm 1.09$ & $11.60 \pm 0.70$ & $14.89 \pm 0.88$ & \underline{$4.17 \pm 0.27$} & $10.19$ \\
EFTL & $6.17 \pm 0.36$ & $9.09 \pm 0.51$ & $6.46 \pm 0.37$ & $18.52 \pm 1.00$ & $11.79 \pm 0.65$ & $15.33 \pm 0.84$ & $4.21 \pm 0.25$ & $10.22$ \\
\midrule
CP-JDOT & $\mathbf{5.59 \pm 0.33}$ & $\mathbf{8.33 \pm 0.48}$ & $\mathbf{6.17 \pm 0.36}$ & $\mathbf{17.54 \pm 0.96}$ & $\mathbf{10.51 \pm 0.60}$ & $\mathbf{13.98 \pm 0.78}$ & $\mathbf{3.99 \pm 0.24}$ & $\mathbf{9.44}$ \\
\bottomrule
\end{tabular}
}
\caption{Avg. Set Size on DomainNet126 (ResNet-50 / 3-Shot / $\alpha = 0.05$ / HPS)}
\label{tab:domainnet126_3shot_005_hps}
\end{table*}

\begin{table*}[t!]
\centering
\resizebox{\textwidth}{!}{
\begin{tabular}{lcccccccc}
\toprule
Method
& R$\rightarrow$C & R$\rightarrow$P & P$\rightarrow$C
& C$\rightarrow$S & S$\rightarrow$P & R$\rightarrow$S & P$\rightarrow$R
& Avg \\
\midrule
S\&T & $20.25 \pm 1.21$ & $24.03 \pm 1.42$ & $20.17 \pm 1.20$ & $39.22 \pm 2.28$ & $28.34 \pm 1.67$ & $32.83 \pm 1.92$ & $14.41 \pm 0.87$ & $25.61$ \\
\midrule
MME & $18.99 \pm 1.10$ & $23.57 \pm 1.35$ & $19.73 \pm 1.14$ & $36.71 \pm 2.06$ & $27.57 \pm 1.57$ & $33.05 \pm 1.87$ & $14.37 \pm 0.84$ & $24.86$ \\
CDAC & $16.93 \pm 0.93$ & $21.56 \pm 1.17$ & $17.55 \pm 0.96$ & $32.31 \pm 1.72$ & $24.80 \pm 1.34$ & $28.13 \pm 1.51$ & $13.53 \pm 0.75$ & $22.12$ \\
DECOTA & $16.80 \pm 0.91$ & $21.59 \pm 1.16$ & $17.73 \pm 0.96$ & $32.99 \pm 1.74$ & $25.10 \pm 1.34$ & $28.44 \pm 1.51$ & $13.63 \pm 0.75$ & $22.33$ \\
IDMNE & \underline{$15.93 \pm 0.82$} & \underline{$20.11 \pm 1.02$} & $16.55 \pm 0.85$ & \underline{$29.89 \pm 1.50$} & \underline{$23.05 \pm 1.17$} & \underline{$26.35 \pm 1.33$} & $12.79 \pm 0.66$ & \underline{$20.67$} \\
ProML & $16.38 \pm 0.85$ & $20.63 \pm 1.06$ & $17.20 \pm 0.89$ & $30.60 \pm 1.55$ & $23.52 \pm 1.20$ & $27.05 \pm 1.37$ & $12.71 \pm 0.67$ & $21.16$ \\
EFTL & $16.05 \pm 0.77$ & $20.32 \pm 0.97$ & \underline{$16.44 \pm 0.79$} & $30.05 \pm 1.41$ & $23.68 \pm 1.12$ & $27.59 \pm 1.30$ & \underline{$12.69 \pm 0.62$} & $20.97$ \\
\midrule
CP-JDOT & $\mathbf{13.85 \pm 0.69}$ & $\mathbf{18.26 \pm 0.90}$ & $\mathbf{15.01 \pm 0.75}$ & $\mathbf{27.45 \pm 1.33}$ & $\mathbf{20.60 \pm 1.01}$ & $\mathbf{25.07 \pm 1.22}$ & $\mathbf{11.54 \pm 0.58}$ & $\mathbf{18.83}$ \\
\bottomrule
\end{tabular}
}
\caption{Avg. Set Size on DomainNet126 (ResNet-50 / 3-Shot / $\alpha = 0.05$ / APS)}
\label{tab:domainnet126_3shot_005_aps}
\end{table*}

\begin{table*}[t!]
\centering
\resizebox{.8\textwidth}{!}{
\begin{tabular}{lcccccc}
\toprule
Method
& HAM$\rightarrow$PH2 & HAM$\rightarrow$MSK & HAM$\rightarrow$Derm7pt
& HAM$\rightarrow$SONIC & HAM$\rightarrow$UDA & Avg \\
\midrule
S\&T & $2.04 \pm 0.32$ & $2.89 \pm 0.45$ & $3.28 \pm 0.40$ & $1.13 \pm 0.17$ & $3.10 \pm 0.49$ & $2.49$ \\
\midrule
MME & $1.66 \pm 0.26$ & $2.45 \pm 0.38$ & $2.86 \pm 0.34$ & \underline{$0.92 \pm 0.01$} & $2.69 \pm 0.43$ & $2.12$ \\
CDAC & $1.59 \pm 0.24$ & $2.36 \pm 0.36$ & $2.74 \pm 0.32$ & $\mathbf{0.91 \pm 0.01}$ & $2.58 \pm 0.41$ & $2.04$ \\
DECOTA & $1.36 \pm 0.19$ & $2.22 \pm 0.34$ & $2.55 \pm 0.30$ & $\mathbf{0.91 \pm 0.01}$ & $2.30 \pm 0.38$ & $1.87$ \\
IDMNE & $1.42 \pm 0.21$ & $1.91 \pm 0.30$ & $2.24 \pm 0.29$ & $\mathbf{0.91 \pm 0.01}$ & $2.19 \pm 0.37$ & $1.73$ \\
ProML & \underline{$1.25 \pm 0.17$} & $1.97 \pm 0.32$ & $2.37 \pm 0.31$ & $\mathbf{0.91 \pm 0.01}$ & $1.94 \pm 0.32$ & $1.69$ \\
EFTL & $1.26 \pm 0.18$ & \underline{$1.77 \pm 0.28$} & \underline{$2.10 \pm 0.26$} & $\mathbf{0.91 \pm 0.01}$ & \underline{$1.87 \pm 0.31$} & \underline{$1.58$} \\
\midrule
CP-JDOT & $\mathbf{1.21 \pm 0.16}$ & $\mathbf{1.69 \pm 0.26}$ & $\mathbf{2.02 \pm 0.24}$ & $\mathbf{0.91 \pm 0.01}$ & $\mathbf{1.76 \pm 0.29}$ & $\mathbf{1.52}$ \\
\bottomrule
\end{tabular}
}
\caption{Avg. Set Size on Skin Lesion Classification (ResNet-18 / 10-Shot / $\alpha = 0.1$ / HPS)}
\label{tab:skin_hps}
\end{table*}

\begin{table*}[t!]
\centering
\resizebox{.8\textwidth}{!}{
\begin{tabular}{lcccccc}
\toprule
Method
& HAM$\rightarrow$PH2 & HAM$\rightarrow$MSK & HAM$\rightarrow$Derm7pt
& HAM$\rightarrow$SONIC & HAM$\rightarrow$UDA & Avg \\
\midrule
S\&T    & $2.87 \pm 0.43$ & $3.98 \pm 0.61$ & $4.54 \pm 0.55$ & $2.23 \pm 0.22$ & $4.37 \pm 0.68$ & $3.60$ \\
\midrule
MME     & $2.43 \pm 0.38$ & $3.54 \pm 0.54$ & $4.12 \pm 0.48$ & $1.17 \pm 0.15$ & $3.93 \pm 0.61$ & $3.04$ \\
CDAC    & $2.34 \pm 0.35$ & $3.43 \pm 0.51$ & $3.98 \pm 0.45$ & $1.26 \pm 0.13$ & $3.82 \pm 0.58$ & $2.97$ \\
DECOTA  & $1.98 \pm 0.31$ & $3.23 \pm 0.48$ & $3.76 \pm 0.42$ & $1.14 \pm 0.20$ & $3.37 \pm 0.54$ & $2.70$ \\
IDMNE  & $2.12 \pm 0.33$ & $2.98 \pm 0.45$ & $3.54 \pm 0.40$ & \underline{$1.12 \pm 0.07$} & $3.62 \pm 0.56$ & $2.68$ \\
ProML    & \underline{$1.87 \pm 0.29$} & $3.12 \pm 0.47$ & $3.87 \pm 0.43$ & \underline{$1.12 \pm 0.15$} & $3.15 \pm 0.51$ & $2.63$ \\
EFTL   & $2.03 \pm 0.30$ & \underline{$2.87 \pm 0.44$} & \underline{$3.43 \pm 0.38$} & $\mathbf{1.08 \pm 0.13}$ & \underline{$3.04 \pm 0.49$} & \underline{$2.49$} \\
\midrule
CP-JDOT    & $\mathbf{1.76 \pm 0.27}$ & $\mathbf{2.65 \pm 0.41}$ & $\mathbf{3.21 \pm 0.35}$ & $1.15 \pm 0.05$ & $\mathbf{2.83 \pm 0.46}$ & $\mathbf{2.32}$ \\
\bottomrule
\end{tabular}
}
\caption{Avg. Set Size on Skin Lesion Classification (ResNet-18 / 10-Shot / $\alpha = 0.1$ / APS) }
\label{tab:skin_aps}
\end{table*}

Table~\ref{tab:skin_hps} reports results on the five skin lesion target domains. The improvement over the baselines is consistent across 4 target datasets (PH2, MSK, Derm7pt, UDA) with the exception of SONIC, where performance, especially when evaluating with the HPS nonconformity score is near-optimal for every algorithm. SONIC has the following distinct characteristic in our experimental setup: Even when we have access to the minimum amount of labeled target samples required to successfully run conformal prediction with safety parameter $\alpha = 0.1$, all baselines apart from S\&T can achieve accuracy higher than $1-\alpha$, in which case simply outputting the most probable class already satisfies the coverage constraint. Thus, especially when evaluating with the HPS score, all methods produce mostly singleton and empty sets and their performance cannot be distinguished. For the rest of the datasets, we see improvements of 4.5\% for HPS and 8\% for APS on average, indicating that incorporating CRM during training can improve the downstream performance of a conformal predictor even in a challenging medical task.

\subsection{Ablation Study for the Conformal Objective} \label{exp:ablation}

In this section, we perform an ablation study on the conformal objectives of CP-JDOT, using two 3-shot Office-Home tasks, $A \to R$, an easier task where the Optimal Transport component works well, and $P \to C$, a challenging task where the Optimal Transport plan induces negative transfer more often. We set $\alpha = 0.05$. The first variant replaces $\mathcal{L}_{\text{cls}}$ with categorical cross-entropy and sets $\lambda_c = 0$, which essentially recovers GOT in a SSDA setting. Here we do not include conformal prediction anywhere in the training pipeline. As such, we group the labeled target data with the labeled source data. The second variant is CP-JDOT with $\lambda_c = 0$. This objective encourages the conformal set of a target sample to include the transported source label but does not push the set to become smaller. The third variant is CP-JDOT with $\lambda_{\text{cls}} = 0$. This objective does not use the transported source labels at all, and only applies the set reduction penalty to the conformal set of a target sample. Finally, we compare the above against the full implementation of CP-JDOT as described in Section \ref{section:method}. 

Table~\ref{tab:ablation_cpjdot} reports the results of our ablation study. Across both tasks, CP-JDOT achieves the smallest average prediction sets at the same reported coverage of 0.95. Relative to GOT with cross-entropy, CP-JDOT reduces average set size by approximately 9--11\% with HPS and 14--16\% with APS, despite having 1.5 and 2.9 percentage points lower top-1 accuracy on $A \to R$ and $P \to C$, respectively. Removing either loss component increases average set size on both tasks under both scores, indicating that anchoring on optimally transported source labels and including a set size objective are both necessary components of CP-JDOT.

\begin{table}[t!]
\centering
\begin{tabular}{llccccc}
\toprule
& & \multicolumn{2}{c}{HPS, $\alpha = 0.05$} & \multicolumn{2}{c}{APS, $\alpha = 0.05$} & \\
\cmidrule(lr){3-4} \cmidrule(lr){5-6}
Task & Variant
& Set size & Coverage
& Set size & Coverage
& Acc.\ (\%) \\
\midrule
$A \rightarrow R$ & CE instead of $\mathcal{L}_{\text{cls}}, \, \lambda_c = 0$ & \underline{$3.57 \pm 0.27$} & 0.95 & \underline{$8.50 \pm 0.54$} & 0.95 & \textbf{81.4} \\
          & $\lambda_c = 0$ & $3.96 \pm 0.31$ & 0.95 & $9.59 \pm 0.60$ & 0.95 & 79.5 \\
          & $\lambda_{\text{cls}} = 0$ & $4.21 \pm 0.35$ & 0.95 & $10.12 \pm 0.67$ & 0.95 & 76.4 \\
          & CP-JDOT & $\mathbf{3.17 \pm 0.22}$ & 0.95 & $\mathbf{7.11 \pm 0.40}$ & 0.95 & \underline{79.9} \\
\addlinespace
$P \rightarrow C$ & CE instead of $\mathcal{L}_{\text{cls}}, \, \lambda_c = 0$ & \underline{$17.38 \pm 1.02$} & 0.95 & \underline{$23.11 \pm 1.09$} & 0.95 & \textbf{62.0} \\
          & $\lambda_c = 0$ & $18.52 \pm 1.16$ & 0.95 & $23.94 \pm 1.21$ & 0.95 & 58.9 \\
          & $\lambda_{\text{cls}} = 0$ & $19.34 \pm 1.30$ & 0.95 & $25.16 \pm 1.38$ & 0.95 & 55.3 \\
          & CP-JDOT & $\mathbf{15.75 \pm 0.78}$ & 0.95 & $\mathbf{19.89 \pm 0.68}$ & 0.95 & \underline{59.1} \\
\bottomrule
\end{tabular}
\caption{Ablation study of the conformal objectives on the 3-shot Office-Home $A \to R$ and $P \to C$ transfer tasks. CP-JDOT achieves the smallest average set size despite performing worse than GOT in terms of top-1 accuracy. Both variants that remove a component perform worse than GOT and full CP-JDOT, indicating that they are necessary to achieve competitive performance.}
\label{tab:ablation_cpjdot}
\end{table}

\subsection{Ablation Study for Set Control} \label{exp:control}

\begin{figure}[t!]
    \centering
    \includegraphics[width=\linewidth]{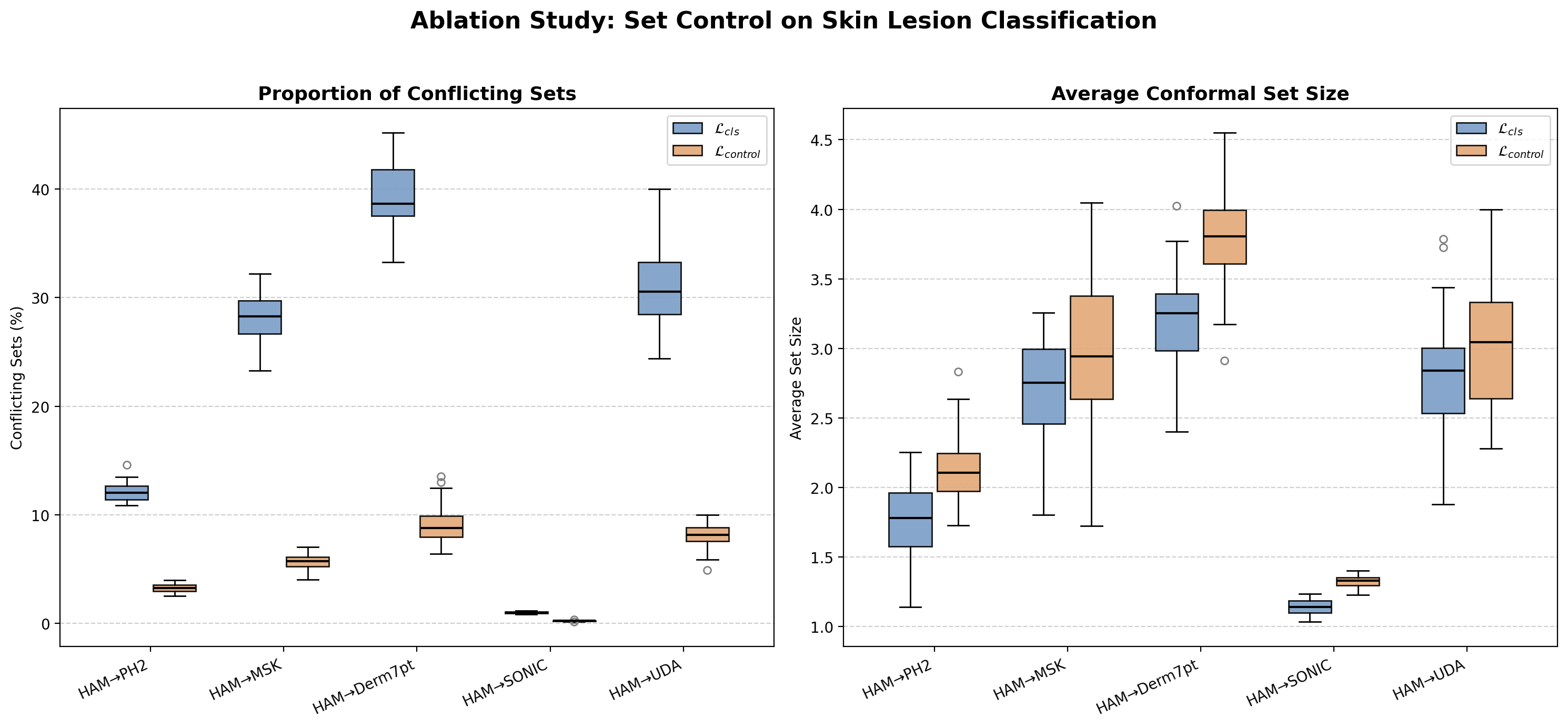}
    \caption{Ablation study comparing CP-JDOT trained with $\mathcal{L}_{\text{cls}}$ versus $\mathcal{L}_{\text{control}}$ on the skin lesion classification setting, evaluating with the APS nonconformity score and $\alpha=0.1$. Training with $\mathcal{L}_{\text{control}}$ consistently reduces the proportion of conflicting sets across all target domains, at the cost of a slight increase in average set size.}
    \label{fig:control}
\end{figure}

In this section, we demonstrate the ability of $\mathcal{L}_{\text{control}}$ in \eqref{eq:special-loss} to shape the composition of conformal sets when domain knowledge is available, using the skin lesion classification task as our case study. Specifically, we consider the clinically motivated constraint that prediction sets should not simultaneously include both benign and malignant classes, as such sets provide conflicting diagnostic signals and reduce clinical utility.

We conduct an ablation study comparing (i) CP-JDOT trained with $\mathcal{L}_{\text{cls}}$, and (ii) CP-JDOT trained with $\mathcal{L}_{\text{control}}$ replacing $\mathcal{L}_{\text{cls}}$, evaluating both with the APS nonconformity score at inference at miscoverage level $\alpha=0.1$. For configuration (ii), $\mathcal{Y}_{\text{err}}$ is defined as the set of classes belonging to the opposite diagnostic category from the transported label $y_i^s$. We evaluate each configuration on two metrics: the proportion of conformal sets containing both benign and malignant classes (conflicting sets), and the average set size.

Figure~\ref{fig:control} shows that training with $\mathcal{L}_{\text{cls}}$ produces conflicting sets in approximately 12\%, 28\%, 38\%, 1\%, and 30\% of the cases in PH2, MSK, Derm7pt, SONIC and UDA respectively. Training with $\mathcal{L}_{\text{control}}$ reduces conflicting sets to 4\%, 6\%, 9\%, 0.5\%, and 9\% respectively, at the cost of a modest increase in average set size. This tradeoff enables practitioners to choose whether they prefer a larger but semantically coherent set than a compact but contradictory one, leading to more actionable outputs. Our ablation study shows that for a 10-20\% increase in average set size, we can reduce the number of conflicting sets by 50-80\% across datasets, significantly decreasing the number of sets that include both benign and malignant labels.

\subsection{Results for different values of \texorpdfstring{$\alpha$}{alpha} at test-time} \label{exp:alpha}

In this section, we present results for two additional test-time miscoverage levels, $\alpha \in \{0.025, 0.1\}$, while keeping the training-time value fixed at $\alpha = 0.05$, for the Office-Home 3-shot setting. Additional results for the 1-shot setting are deferred to Appendix \ref{additional:alpha}. The importance of this experiment stems from the fact that baseline methods use conformal prediction only at the post-hoc calibration step, so their target coverage level can be chosen after training. Our method requires choosing $\alpha$ during training, when the coverage level that will be chosen at deployment may still be unknown. The aim of this experiment is therefore to assess whether a single model trained at $\alpha = 0.05$ retains its set size gains compared to the baselines even at different test-time miscoverage levels, as well as to derive practical tips on how to choose training-time $\alpha$ when test-time miscoverage is unknown a priori. Tables~\ref{tab:office_home_3shot_0025_hps}, \ref{tab:office_home_3shot_01_hps}, \ref{tab:office_home_3shot_0025_aps}, and \ref{tab:office_home_3shot_01_aps} present our experimental results. Our algorithm still consistently improves over the best baseline, with the improvements being slightly lower than the setting where the training-time $\alpha$ matches the test-time $\alpha$. The reduction in improvement is more pronounced when test-time $\alpha=0.025$, where the deployment quantile is higher than the training quantile. On the other hand, when test-time $\alpha=0.1$, the improvement is closer to the matched $\alpha$ setting. This asymmetry appears in other related works~\citep{stutz2021learning, marani2026classadaptiveconformaltraining}, indicating that, for practitioners, training with a slightly smaller $\alpha$ than the one anticipated at deployment may be preferable.



\begin{table*}[t!]
\centering
\resizebox{\textwidth}{!}{
\begin{tabular}{lccccccccccccc}
\toprule
Method
& A$\rightarrow$C & A$\rightarrow$P & A$\rightarrow$R
& C$\rightarrow$A & C$\rightarrow$P & C$\rightarrow$R
& P$\rightarrow$A & P$\rightarrow$C & P$\rightarrow$R
& R$\rightarrow$A & R$\rightarrow$C & R$\rightarrow$P
& Avg \\
\midrule
S\&T & $27.16 \pm 1.62$ & $9.10 \pm 0.71$ & $7.09 \pm 0.54$ & $19.68 \pm 1.57$ & $12.33 \pm 1.14$ & $9.76 \pm 0.80$ & $23.75 \pm 2.11$ & $28.98 \pm 1.53$ & $6.71 \pm 0.43$ & $16.06 \pm 2.29$ & $25.51 \pm 1.53$ & $6.58 \pm 0.59$ & $16.06$ \\
\midrule
MME & $25.95 \pm 1.48$ & $8.52 \pm 0.64$ & $6.61 \pm 0.48$ & $18.73 \pm 1.43$ & $11.71 \pm 1.02$ & $9.13 \pm 0.71$ & $22.89 \pm 1.90$ & $27.68 \pm 1.37$ & $6.24 \pm 0.38$ & $15.14 \pm 2.12$ & $24.31 \pm 1.45$ & $6.12 \pm 0.54$ & $15.25$ \\
CDAC & $25.51 \pm 1.51$ & $8.16 \pm 0.57$ & $6.19 \pm 0.41$ & $18.24 \pm 1.34$ & $11.16 \pm 0.96$ & $8.85 \pm 0.64$ & $22.37 \pm 1.86$ & $26.81 \pm 1.26$ & $6.04 \pm 0.34$ & $14.86 \pm 1.91$ & $23.66 \pm 1.32$ & $5.74 \pm 0.47$ & $14.80$ \\
DECOTA & $25.64 \pm 1.43$ & $7.99 \pm 0.59$ & $6.34 \pm 0.43$ & $17.92 \pm 1.42$ & $11.31 \pm 0.98$ & $8.61 \pm 0.60$ & $21.62 \pm 1.80$ & $27.02 \pm 1.19$ & $5.91 \pm 0.37$ & $14.72 \pm 2.04$ & $23.91 \pm 1.29$ & $5.62 \pm 0.49$ & $14.72$ \\
IDMNE & $23.79 \pm 1.32$ & $7.67 \pm 0.51$ & $5.82 \pm 0.37$ & $17.41 \pm 1.30$ & $10.34 \pm 0.94$ & $8.08 \pm 0.55$ & $20.12 \pm 1.76$ & $25.18 \pm 1.03$ & $5.57 \pm 0.32$ & $13.78 \pm 1.82$ & $22.14 \pm 1.11$ & $5.39 \pm 0.43$ & $13.77$ \\
ProML & $24.52 \pm 1.40$ & $7.39 \pm 0.46$ & $5.67 \pm 0.34$ & $17.58 \pm 1.24$ & $10.11 \pm 0.88$ & $8.39 \pm 0.61$ & $20.94 \pm 1.84$ & $25.86 \pm 1.08$ & $5.66 \pm 0.31$ & $13.59 \pm 2.06$ & $22.68 \pm 1.18$ & $5.16 \pm 0.40$ & $13.96$ \\
EFTL & \underline{$23.11 \pm 1.29$} & \underline{$7.34 \pm 0.43$} & \underline{$5.54 \pm 0.33$} & \underline{$16.62 \pm 1.20$} & \underline{$9.67 \pm 0.85$} & \underline{$7.69 \pm 0.52$} & \underline{$19.98 \pm 1.72$} & $\mathbf{24.01 \pm 0.87}$ & \underline{$5.42 \pm 0.29$} & \underline{$13.22 \pm 1.96$} & \underline{$21.26 \pm 1.04$} & \underline{$5.03 \pm 0.37$} & \underline{$13.24$} \\
\midrule
CP-JDOT & $\mathbf{22.60 \pm 1.36}$ & $\mathbf{6.94 \pm 0.40}$ & $\mathbf{5.23 \pm 0.30}$ & $\mathbf{16.09 \pm 1.16}$ & $\mathbf{9.24 \pm 0.84}$ & $\mathbf{7.25 \pm 0.50}$ & $\mathbf{19.27 \pm 1.67}$ & \underline{$24.18 \pm 0.78$} & $\mathbf{5.13 \pm 0.27}$ & $\mathbf{12.77 \pm 2.18}$ & $\mathbf{20.59 \pm 0.96}$ & $\mathbf{4.78 \pm 0.36}$ & $\mathbf{12.84}$ \\
\bottomrule
\end{tabular}
}
\caption{Avg. Set Size on Office-Home (ResNet-34 / 3-Shot / $\alpha = 0.025$ / HPS)}
\label{tab:office_home_3shot_0025_hps}
\end{table*}

\begin{table*}[t!]
\centering
\resizebox{\textwidth}{!}{
\begin{tabular}{lccccccccccccc}
\toprule
Method
& A$\rightarrow$C & A$\rightarrow$P & A$\rightarrow$R
& C$\rightarrow$A & C$\rightarrow$P & C$\rightarrow$R
& P$\rightarrow$A & P$\rightarrow$C & P$\rightarrow$R
& R$\rightarrow$A & R$\rightarrow$C & R$\rightarrow$P
& Avg \\
\midrule
S\&T & $9.01 \pm 0.64$ & $2.38 \pm 0.15$ & $2.25 \pm 0.14$ & $5.62 \pm 0.36$ & $2.96 \pm 0.24$ & $2.71 \pm 0.17$ & $7.04 \pm 0.64$ & $10.79 \pm 0.88$ & $2.01 \pm 0.12$ & $4.22 \pm 0.34$ & $8.08 \pm 0.55$ & $1.79 \pm 0.10$ & $4.91$ \\
\midrule
MME & $8.54 \pm 0.56$ & $2.22 \pm 0.12$ & $2.10 \pm 0.11$ & $5.28 \pm 0.33$ & $2.79 \pm 0.21$ & $2.58 \pm 0.14$ & $6.65 \pm 0.57$ & $10.15 \pm 0.77$ & $1.93 \pm 0.10$ & $4.01 \pm 0.29$ & $7.80 \pm 0.49$ & $1.63 \pm 0.08$ & $4.64$ \\
CDAC & $8.30 \pm 0.51$ & $2.16 \pm 0.09$ & $2.01 \pm 0.10$ & $5.16 \pm 0.29$ & $2.64 \pm 0.17$ & $2.46 \pm 0.12$ & $6.36 \pm 0.52$ & $9.82 \pm 0.72$ & $1.84 \pm 0.08$ & $3.81 \pm 0.24$ & $7.53 \pm 0.45$ & $1.57 \pm 0.06$ & $4.47$ \\
DECOTA & $8.12 \pm 0.53$ & $2.09 \pm 0.10$ & $2.04 \pm 0.09$ & $5.03 \pm 0.31$ & $2.69 \pm 0.16$ & $2.42 \pm 0.10$ & $6.23 \pm 0.49$ & $9.61 \pm 0.68$ & $1.88 \pm 0.08$ & $3.84 \pm 0.22$ & $7.46 \pm 0.42$ & $1.56 \pm 0.07$ & $4.41$ \\
IDMNE & $7.83 \pm 0.48$ & $2.01 \pm 0.07$ & $1.92 \pm 0.07$ & $4.84 \pm 0.27$ & $2.55 \pm 0.14$ & $2.30 \pm 0.08$ & $5.99 \pm 0.46$ & $9.23 \pm 0.63$ & $1.76 \pm 0.06$ & $3.61 \pm 0.19$ & $7.10 \pm 0.36$ & $1.49 \pm 0.05$ & $4.22$ \\
ProML & $7.99 \pm 0.49$ & $1.98 \pm 0.06$ & \underline{$1.86 \pm 0.06$} & $4.96 \pm 0.25$ & $2.49 \pm 0.13$ & $2.36 \pm 0.09$ & $6.14 \pm 0.50$ & $9.08 \pm 0.65$ & $1.81 \pm 0.07$ & $3.67 \pm 0.21$ & $7.25 \pm 0.40$ & $1.45 \pm 0.04$ & $4.25$ \\
EFTL & \underline{$7.63 \pm 0.43$} & \underline{$1.96 \pm 0.05$} & $1.87 \pm 0.06$ & \underline{$4.72 \pm 0.24$} & \underline{$2.41 \pm 0.12$} & \underline{$2.28 \pm 0.07$} & \underline{$5.91 \pm 0.44$} & \underline{$8.87 \pm 0.59$} & \underline{$1.71 \pm 0.06$} & \underline{$3.49 \pm 0.17$} & \underline{$6.96 \pm 0.33$} & $\mathbf{1.32 \pm 0.03}$ & \underline{$4.09$} \\
\midrule
CP-JDOT & $\mathbf{7.42 \pm 0.45}$ & $\mathbf{1.86 \pm 0.04}$ & $\mathbf{1.76 \pm 0.05}$ & $\mathbf{4.54 \pm 0.23}$ & $\mathbf{2.29 \pm 0.11}$ & $\mathbf{2.16 \pm 0.06}$ & $\mathbf{5.67 \pm 0.41}$ & $\mathbf{8.62 \pm 0.57}$ & $\mathbf{1.63 \pm 0.05}$ & $\mathbf{3.32 \pm 0.15}$ & $\mathbf{6.71 \pm 0.30}$ & \underline{$1.35 \pm 0.02$} & $\mathbf{3.94}$ \\
\bottomrule
\end{tabular}
}
\caption{Avg. Set Size on Office-Home (ResNet-34 / 3-Shot / $\alpha = 0.1$ / HPS)}
\label{tab:office_home_3shot_01_hps}
\end{table*}


\begin{table*}[t!]
\centering
\resizebox{\textwidth}{!}{
\begin{tabular}{lccccccccccccc}
\toprule
Method
& A$\rightarrow$C & A$\rightarrow$P & A$\rightarrow$R
& C$\rightarrow$A & C$\rightarrow$P & C$\rightarrow$R
& P$\rightarrow$A & P$\rightarrow$C & P$\rightarrow$R
& R$\rightarrow$A & R$\rightarrow$C & R$\rightarrow$P
& Avg \\
\midrule
S\&T & $33.72 \pm 1.64$ & $18.31 \pm 1.03$ & $14.54 \pm 1.12$ & $25.59 \pm 1.70$ & $20.14 \pm 1.29$ & $16.30 \pm 0.92$ & $30.28 \pm 1.97$ & $33.78 \pm 1.65$ & $14.97 \pm 0.78$ & $23.63 \pm 1.96$ & $30.46 \pm 1.71$ & $14.28 \pm 1.03$ & $23.00$ \\
\midrule
MME & $32.27 \pm 1.48$ & $17.02 \pm 0.91$ & $13.31 \pm 1.02$ & $24.48 \pm 1.53$ & $18.92 \pm 1.12$ & $15.27 \pm 0.79$ & $29.17 \pm 1.78$ & $32.36 \pm 1.43$ & $13.81 \pm 0.68$ & $22.48 \pm 1.83$ & $29.01 \pm 1.56$ & $13.21 \pm 0.91$ & $21.78$ \\
CDAC & $31.22 \pm 1.37$ & $16.58 \pm 0.81$ & $12.89 \pm 0.91$ & $23.69 \pm 1.41$ & $18.29 \pm 1.03$ & $14.72 \pm 0.72$ & $28.45 \pm 1.69$ & $31.51 \pm 1.28$ & $13.22 \pm 0.60$ & $21.62 \pm 1.67$ & $28.13 \pm 1.44$ & $12.52 \pm 0.83$ & $21.07$ \\
DECOTA & $31.57 \pm 1.40$ & $16.15 \pm 0.74$ & $12.97 \pm 0.94$ & $23.33 \pm 1.37$ & $17.91 \pm 0.96$ & $14.91 \pm 0.76$ & $27.79 \pm 1.60$ & $31.80 \pm 1.32$ & $13.38 \pm 0.63$ & $21.36 \pm 1.72$ & $27.78 \pm 1.39$ & $12.61 \pm 0.80$ & $20.96$ \\
IDMNE & $29.43 \pm 1.24$ & $15.32 \pm 0.65$ & $12.01 \pm 0.82$ & $22.24 \pm 1.25$ & $17.09 \pm 0.88$ & $13.63 \pm 0.63$ & $26.36 \pm 1.48$ & $29.85 \pm 1.09$ & $12.31 \pm 0.50$ & $20.48 \pm 1.57$ & $26.48 \pm 1.25$ & $11.85 \pm 0.72$ & $19.75$ \\
ProML & $30.06 \pm 1.29$ & $14.97 \pm 0.59$ & $11.77 \pm 0.79$ & $22.58 \pm 1.31$ & $16.62 \pm 0.80$ & $13.89 \pm 0.67$ & $26.95 \pm 1.55$ & $30.42 \pm 1.15$ & $12.47 \pm 0.53$ & $20.73 \pm 1.54$ & $26.82 \pm 1.30$ & $11.22 \pm 0.66$ & $19.88$ \\
EFTL & \underline{$28.71 \pm 1.17$} & \underline{$14.77 \pm 0.54$} & \underline{$11.65 \pm 0.76$} & \underline{$21.49 \pm 1.15$} & \underline{$16.20 \pm 0.76$} & \underline{$13.19 \pm 0.59$} & \underline{$25.78 \pm 1.40$} & $\mathbf{27.35 \pm 0.88}$ & \underline{$11.60 \pm 0.44$} & \underline{$19.80 \pm 1.46$} & \underline{$25.71 \pm 1.17$} & \underline{$10.82 \pm 0.63$} & \underline{$18.92$} \\
\midrule
CP-JDOT & $\mathbf{26.89 \pm 1.09}$ & $\mathbf{13.36 \pm 0.44}$ & $\mathbf{10.47 \pm 0.73}$ & $\mathbf{20.03 \pm 1.09}$ & $\mathbf{14.98 \pm 0.70}$ & $\mathbf{12.00 \pm 0.55}$ & $\mathbf{24.06 \pm 1.33}$ & \underline{$27.57 \pm 0.73$} & $\mathbf{10.42 \pm 0.36}$ & $\mathbf{18.19 \pm 1.43}$ & $\mathbf{24.12 \pm 1.09}$ & $\mathbf{9.71 \pm 0.56}$ & $\mathbf{17.65}$ \\
\bottomrule
\end{tabular}
}
\caption{Avg. Set Size on Office-Home (ResNet-34 / 3-Shot / $\alpha = 0.025$ / APS)}
\label{tab:office_home_3shot_0025_aps}
\end{table*}

\begin{table*}[t!]
\centering
\resizebox{\textwidth}{!}{
\begin{tabular}{lccccccccccccc}
\toprule
Method
& A$\rightarrow$C & A$\rightarrow$P & A$\rightarrow$R
& C$\rightarrow$A & C$\rightarrow$P & C$\rightarrow$R
& P$\rightarrow$A & P$\rightarrow$C & P$\rightarrow$R
& R$\rightarrow$A & R$\rightarrow$C & R$\rightarrow$P
& Avg \\
\midrule
S\&T & $15.13 \pm 0.88$ & $7.24 \pm 0.42$ & $6.16 \pm 0.32$ & $10.94 \pm 0.81$ & $8.41 \pm 0.62$ & $7.12 \pm 0.36$ & $12.74 \pm 0.98$ & $16.27 \pm 0.96$ & $5.92 \pm 0.37$ & $9.32 \pm 0.74$ & $13.75 \pm 0.68$ & $5.07 \pm 0.31$ & $9.84$ \\
\midrule
MME & $14.23 \pm 0.79$ & $6.83 \pm 0.36$ & $5.85 \pm 0.27$ & $10.47 \pm 0.75$ & $7.94 \pm 0.52$ & $6.75 \pm 0.30$ & $12.25 \pm 0.89$ & $15.32 \pm 0.85$ & $5.61 \pm 0.33$ & $8.73 \pm 0.67$ & $13.02 \pm 0.59$ & $4.76 \pm 0.27$ & $9.31$ \\
CDAC & $13.78 \pm 0.73$ & $6.59 \pm 0.31$ & $5.55 \pm 0.24$ & $10.08 \pm 0.69$ & $7.51 \pm 0.47$ & $6.54 \pm 0.26$ & $11.99 \pm 0.84$ & $14.90 \pm 0.77$ & $5.28 \pm 0.29$ & $8.41 \pm 0.61$ & $12.67 \pm 0.55$ & $4.57 \pm 0.23$ & $8.99$ \\
DECOTA & $13.52 \pm 0.75$ & $6.41 \pm 0.29$ & $5.68 \pm 0.25$ & $9.94 \pm 0.72$ & $7.28 \pm 0.43$ & $6.33 \pm 0.25$ & $11.72 \pm 0.79$ & $14.64 \pm 0.73$ & $5.37 \pm 0.29$ & $8.46 \pm 0.58$ & $12.42 \pm 0.51$ & $4.63 \pm 0.24$ & $8.87$ \\
IDMNE & $12.69 \pm 0.64$ & $6.00 \pm 0.25$ & $5.15 \pm 0.20$ & $9.55 \pm 0.65$ & $6.93 \pm 0.39$ & $5.94 \pm 0.21$ & $11.17 \pm 0.73$ & $13.93 \pm 0.64$ & $4.89 \pm 0.23$ & $7.81 \pm 0.53$ & $11.83 \pm 0.47$ & $4.26 \pm 0.19$ & $8.35$ \\
ProML & $13.04 \pm 0.69$ & $5.84 \pm 0.22$ & $4.96 \pm 0.17$ & $9.72 \pm 0.67$ & $6.64 \pm 0.36$ & $6.11 \pm 0.23$ & $11.48 \pm 0.77$ & $14.18 \pm 0.68$ & $5.03 \pm 0.25$ & $7.98 \pm 0.56$ & $12.07 \pm 0.49$ & $4.03 \pm 0.16$ & $8.42$ \\
EFTL & \underline{$12.32 \pm 0.61$} & \underline{$5.71 \pm 0.20$} & \underline{$4.82 \pm 0.15$} & \underline{$9.28 \pm 0.59$} & \underline{$6.38 \pm 0.33$} & \underline{$5.76 \pm 0.18$} & $\mathbf{10.29 \pm 0.65}$ & \underline{$13.43 \pm 0.57$} & \underline{$4.58 \pm 0.19$} & \underline{$7.48 \pm 0.49$} & \underline{$11.41 \pm 0.42$} & \underline{$3.94 \pm 0.15$} & \underline{$7.95$} \\
\midrule
CP-JDOT & $\mathbf{11.48 \pm 0.59}$ & $\mathbf{5.01 \pm 0.17}$ & $\mathbf{4.36 \pm 0.12}$ & $\mathbf{8.86 \pm 0.58}$ & $\mathbf{5.83 \pm 0.31}$ & $\mathbf{5.21 \pm 0.15}$ & \underline{$10.36 \pm 0.69$} & $\mathbf{12.88 \pm 0.51}$ & $\mathbf{4.03 \pm 0.16}$ & $\mathbf{6.87 \pm 0.47}$ & $\mathbf{10.44 \pm 0.38}$ & $\mathbf{3.51 \pm 0.12}$ & $\mathbf{7.40}$ \\
\bottomrule
\end{tabular}
}
\caption{Avg. Set Size on Office-Home (ResNet-34 / 3-Shot / $\alpha = 0.1$ / APS)}
\label{tab:office_home_3shot_01_aps}
\end{table*}

\subsection{Results for different combinations of labeled samples per class and \texorpdfstring{$\alpha$}{alpha}} \label{exp:data}

\begin{figure}[t!]
    \centering
    \includegraphics[width=\linewidth]{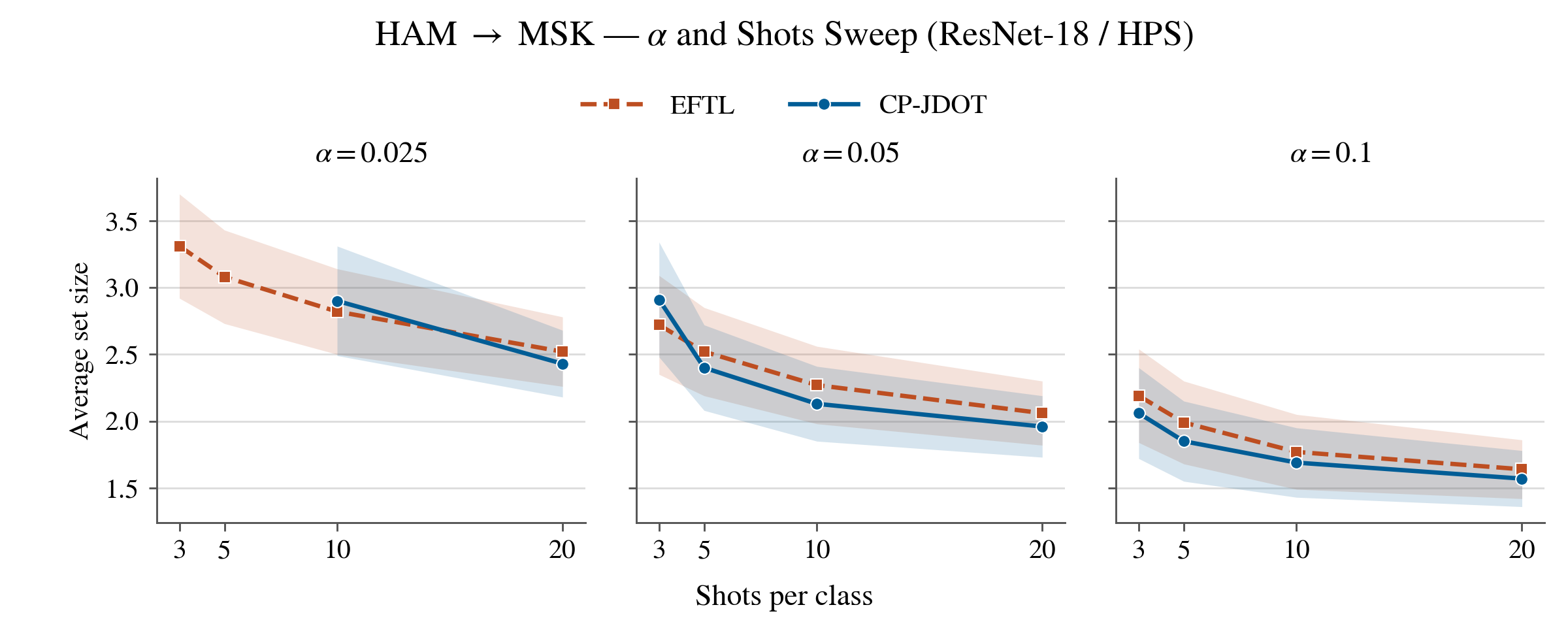}
    \caption{Labeled target data count sweeps for varying levels of miscoverage $\alpha$, evaluated with the HPS score. Our method is infeasible in the 3-shot and 5-shot settings for $\alpha=0.025$ since there are too few labeled target training samples to estimate the training-time quantile, and trails EFTL in the 10-shot setting for $\alpha=0.025$ and in the 3-shot setting for $\alpha=0.05$, where the conformal quantile can be unstable due to few data. In every other setting, we have enough samples to calculate a stable quantile and our method outperforms EFTL.}
    \label{fig:sweep_hps}
\end{figure}

\begin{figure}[t!]
    \centering
    \includegraphics[width=\linewidth]{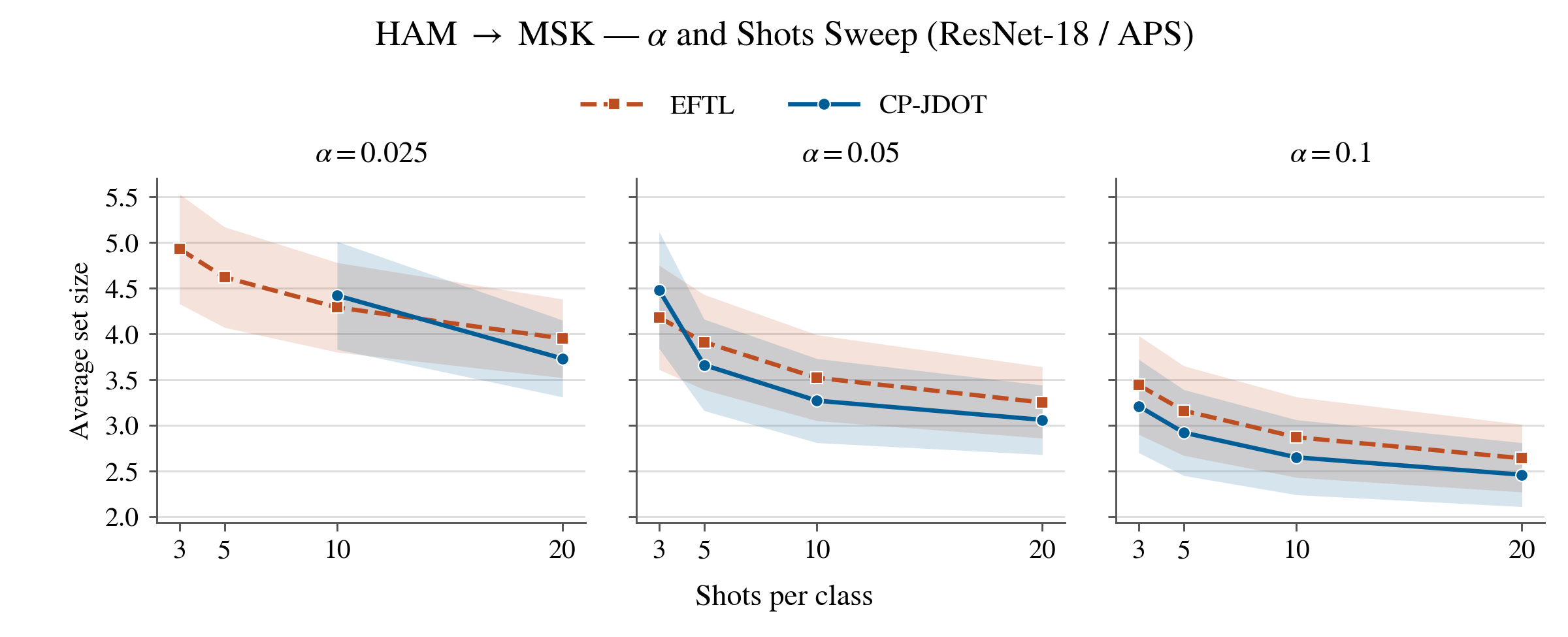}
    \caption{Labeled target data count sweeps for varying levels of miscoverage $\alpha$, evaluated with the APS score. Our method is infeasible in the 3-shot and 5-shot settings for $\alpha=0.025$ since there are too few labeled target training samples to estimate the training-time quantile, and trails EFTL in the 10-shot setting for $\alpha=0.025$ and in the 3-shot setting for $\alpha=0.05$, where the conformal quantile can be unstable due to few data. In every other setting, we have enough samples to calculate a stable quantile and our method outperforms EFTL.}
    \label{fig:sweep_aps}
\end{figure}

In this section, we vary the number of labeled target data per class to assess our algorithm’s sensitivity to label availability, particularly at small values of $\alpha$ where the conformal quantile is more sensitive to the number of samples used to estimate it. We evaluate on HAM$\rightarrow$MSK, where both domains contain all 7 classes, and compare our algorithm to the strongest baseline (EFTL) at different miscoverage levels. We sweep 3, 5, 10, and 20 shots per class at $\alpha \in \{0.025, 0.05, 0.1\}$ and report average prediction set size for HPS and APS in Figures~\ref{fig:sweep_hps} and~\ref{fig:sweep_aps}, respectively. Shaded bands show $\pm 1$ standard deviation across the ten splits. EFTL produces smaller sets as the number of shots increases. Its training does not require estimating a conformal quantile from the few labeled target examples, so it is less sensitive to small target sample sizes. Our method is infeasible at 3 and 5 shots for $\alpha=0.025$ because the minimum total number of samples required to estimate the quantile at training time is $\lceil 1/\alpha \rceil - 1$. It trails EFTL slightly at 10 shots for $\alpha=0.025$ and by a larger margin at 3 shots for $\alpha=0.05$, two settings where the quantile estimate can be unstable. Our method produces smaller sets in every other feasible setting for both scores, with larger gains under APS, and maintains an advantage at 20 shots. For practitioners, these results suggest preferring our method when the total number of labeled target data available during training is comfortably above $1/\alpha$, since the conformal quantile can be unstable when the number of available samples is near that threshold.

%% file: sections/conclusion.tex
We introduced CP-JDOT, a framework for integrating conformal risk minimization into semi-supervised domain adaptation tasks. Unlike the conventional approach that would first optimize a domain-adaptation objective and subsequently apply conformal prediction as a post-hoc calibration step, CP-JDOT incorporates a conformal alignment objective directly into training, allowing the learned representation and prediction scores to better reflect the geometry required for efficient prediction sets. Beyond set-size reduction, CP-JDOT provides a flexible mechanism to enable adding domain-specific structure into the prediction sets through a customizable training loss. Together, these results highlight the potential of jointly optimizing domain adaptation and conformal objectives rather than treating uncertainty quantification as a separate post-processing step. We believe this work can serve as a useful foundation for developing domain-adaptive models outputting prediction sets that are not only statistically calibrated, but also efficient and aligned with application-specific requirements.

%% file: appendix/got.tex
We use the label-guided Geometric Optimal Transport (GOT) cost proposed
by~\cite{maman2025tmlr} as the transport cost in
CP-JDOT. Unlike the squared Euclidean cost used in DeepJDOT, which only
depends on the distance between an individual source--target pair, GOT
incorporates the local geometry of both domains through diffusion
processes. The method further incorporates the available source labels
to guide the diffusion process according to class structure.

Let
\[
    z_i^s = g(x_i^s), \qquad
    z_j^u = g(x_j^u)
\]
denote the source and target representations in a minibatch of size
$B$. GOT first constructs three affinity matrices corresponding to
source-domain, target-domain, and cross-domain relationships. We denote
these matrices by $K^s \in \mathbb{R}^{B\times B}$,
$K^t \in \mathbb{R}^{B\times B}$, and
$K^{st} \in \mathbb{R}^{B\times B}$, respectively. The source affinity
matrix is made label-aware by incorporating the known source labels:
\begin{equation}
    K^s_{ii'}
    =
    k(z_i^s,z_{i'}^s)\,
    \mathbbm{1}\!\left[y_i^s = y_{i'}^s\right],
    \label{eq:got-source-kernel}
\end{equation}
where $k(\cdot,\cdot)$ is the affinity kernel used to measure local
similarity. The target and cross-domain affinity matrices are given by
\begin{align}
    K^t_{jj'} &= k(z_j^u,z_{j'}^u),
    \label{eq:got-target-kernel}\\
    K^{st}_{ij} &= k(z_i^s,z_j^u).
    \label{eq:got-cross-kernel}
\end{align}

The affinity matrices are normalized to obtain diffusion operators for
the source domain, target domain, and source--target transition. We
denote these operators by $P^s$, $P^t$, and $Q$, respectively. GOT
combines these operators into a source-to-target diffusion operator
\begin{equation}
    S = P^s Q P^t,
    \label{eq:got-diffusion}
\end{equation}
whose $(i,j)$-th entry quantifies the diffusion-based connectivity
between source sample $i$ and target sample $j$. This quantity captures
both the geometry within each domain and the relationship between the
two domains.

The GOT transport cost is obtained by taking the negative logarithm of
the diffusion-based transport probability:
\begin{equation}
    C^{\mathrm{GOT}}_{ij}
    =
    -\log S_{ij}.
    \label{eq:got-cost}
\end{equation}
Thus, the transport cost used by CP-JDOT is
\begin{equation}
    c\!\left(g(x_i^s),g(x_j^u)\right)
    =
    C^{\mathrm{GOT}}_{ij},
    \label{eq:got-cost-cp-jdot}
\end{equation}
which replaces the squared Euclidean distance in
Problem~\ref{eq:ot}. The source-label information in
\eqref{eq:got-source-kernel} encourages the resulting diffusion
geometry to respect the class structure of the source domain.

All remaining details of the affinity construction, normalization, and
diffusion process follow~\cite{maman2025tmlr}. We use
the label-guided GOT construction as a drop-in replacement for the
Euclidean transport cost in DeepJDOT.

%% file: appendix/experimental_details.tex
We implement CP-JDOT in PyTorch and conduct all experiments on a workstation
equipped with an NVIDIA TITAN RTX GPU with 24~GB of VRAM and an Intel Core
i9-9980XE CPU running at 3.00~GHz.

\subsection{CP-JDOT Implementation Details}
\label{app:implementation}

\paragraph{Minibatches and training threshold.}
At each iteration, we sample two mini-batches, one from the labeled source
set and one from the shuffled unlabeled target set. We sample the source
mini-batch so that all observed classes are represented equally, using the
largest balanced batch that fits the specified batch size. We also pass
all examples in the few-shot labeled target set $\mathcal{T}_\ell$ through
the current model. For all training images, we first resize them to
$256\times256$ and then apply a random resized crop of size $224\times224$,
random horizontal flipping, and ImageNet normalization.

During training, we use the HPS nonconformity score
$s_\theta(x,y)=1-p_\theta(y\mid x)$, where $p_\theta$ is the softmax output.
To compute the conformal threshold $\tau$, we sort the true-label scores on
$\mathcal{T}_\ell$, denoted by $s_{(1)}\leq\cdots\leq s_{(n_\ell)}$, and
linearly interpolate them at the fractional rank
\[
    r=\min\bigl\{n_\ell,\max\{1,(n_\ell+1)(1-\alpha_{\mathrm{train}})\}\bigr\}.
\]
If $k=\lfloor r\rfloor$ and $m=\lceil r\rceil$, then
$\tau=(1-r+k)s_{(k)}+(r-k)s_{(m)}$.
We use the resulting threshold to define the soft prediction sets on the
unlabeled target mini-batch.

\paragraph{GOT cost and transport coupling.}
To construct the GOT feature cost, we compute squared cosine distances
within each domain and squared Euclidean distances across domains. We
divide each distance matrix by its median before constructing the Gaussian
kernels. For the source kernel, we keep pairs from the same class. For the
target kernel, we keep the four smallest distances in each column. The
cross-domain kernel includes all source--target pairs.

We obtain the diffusion operators $P^s$, $P^t$, and $Q$ using balanced
Sinkhorn scaling. The two within-domain operators have unit row and column
sums. If the source and target mini-batches contain $B_s$ and $B_t$ examples,
respectively, $Q$ has unit row sums and column sums equal to $B_s/B_t$.
The GOT feature cost is then $C^{\mathrm{GOT}}=-\log(P^s Q P^t)$, where the
logarithm is applied elementwise.

For an unlabeled target example $j$, we approximate the inclusion of class
$y$ in its prediction set by
\[
    A_{jy}=\sigma\!\left(\frac{\tau-s_\theta(x_j^u,y)}{T}\right).
\]
Here, $T$ is the sigmoid temperature. The cost of excluding the
transported source label $y_i^s$ from target example $j$ is
$E_{ij}=1-A_{j,y_i^s}$. We combine this with the GOT feature cost to obtain
\[
    M_{ij}=\lambda_a C^{\mathrm{GOT}}_{ij}+\lambda_{\mathrm{cls}}E_{ij}.
\]
We then use the unbalanced Sinkhorn solver in POT~0.9.3 to solve
\begin{equation}
\label{eq:cp-jdot-uot-implementation}
\begin{aligned}
\min_{\gamma\geq 0}\quad &
    \sum_{i,j}\gamma_{ij}M_{ij}
    +\varepsilon_{\mathrm{OT}}\sum_{i,j}\gamma_{ij}(\log\gamma_{ij}-1) \\
    &+\rho\bigl[
        \mathrm{KL}(\gamma\mathbf{1}\,\|\,a)
        +\mathrm{KL}(\gamma^\top\mathbf{1}\,\|\,b)
    \bigr],
\end{aligned}
\end{equation}
where $a$ and $b$ are uniform probability vectors on the two mini-batches
and $\mathrm{KL}$ is the generalized KL divergence.

\paragraph{Model update and optimization.}
Once the coupling has been computed, we hold it fixed and update the model
using one SGD step. The objective consists of the source cross-entropy, the
joint cost weighted by the coupling, and the soft set-size penalty
$\lambda_c B_t^{-1}\sum_{j,y}A_{jy}$. We backpropagate through both the
conformal threshold and the GOT diffusion costs. For the set-control
experiments, we compute the coupling in the same way and use the set-control
loss in place of the transported-label term when updating the model.

For all three datasets, we use SGD with Nesterov momentum $0.9$ and weight decay
$10^{-3}$. At iteration $t$, the backbone learning rate is
$\eta_0(1+0.001t)^{-0.75}$, with base learning rate $\eta_0=0.001$. We use
ten times this learning rate for the linear bottleneck and classifier.
We set $\lambda_c=0.01$ for the set-size penalty and $T=0.01$ for the
sigmoid temperature. For the transport problem, we set the Gaussian scale
to $\epsilon=1$ for all three kernels and use
$\varepsilon_{\mathrm{OT}}=0.02$ and $\rho=0.5$ for the unbalanced
Sinkhorn solve. The released code uses $\alpha_{\mathrm{train}}=0.05$ by
default.

\subsection{Dataset Splits and Conformal Evaluation}
\label{app:cp-jdot-splits-evaluation}

\paragraph{Office-Home.}
For Office-Home, we consider all 12 ordered transfer directions between
Art, Clipart, Product, and Real World. We evaluate both the 1-shot and
3-shot settings, where one and three labeled target training images are
available per class, respectively.

\paragraph{DomainNet126.}
For DomainNet126, we follow the 126-class SSDA setup of
\citet{saito2019semi}, which uses images from Real (R), Clipart (C),
Painting (P), and Sketch (S). We evaluate seven transfer directions:
R$\rightarrow$C, R$\rightarrow$P, P$\rightarrow$C, C$\rightarrow$S,
S$\rightarrow$P, R$\rightarrow$S, and P$\rightarrow$R. We consider
both the 1-shot and 3-shot settings, using one and three labeled target
training images per class, respectively.

\paragraph{Skin lesion classification.}
For the skin lesion experiments, HAM10000 is part of the labeled source
set for every transfer direction. When one of MSK, Derm7pt, SONIC, PH2 and UDA
is used as the target, we combine the other 4 datasets with HAM10000
to form the source set. In the
10-shot setting, we use ten labeled target training images for each class
present in the target domain.

\paragraph{Post-hoc calibration and evaluation.}
For all three benchmarks, we use the remaining images in the target training
split as unlabeled target data. We divide the separate labeled target
holdout into two halves, using one for calibration and the other for
testing. We repeat this procedure for ten random splits using the same
trained model. The released trainer evaluates the final training
checkpoint. When using ten-crop evaluation, we average the softmax
probabilities across the crops before computing the scores.

For each split, we calibrate separately using the HPS and randomized APS
scores. For APS, we sample one uniform random variable per image and use
it for all candidate labels of that image. For either score, we set the
threshold to the
$\lceil(n_{\mathrm{cal}}+1)(1-\alpha_{\mathrm{eval}})\rceil$-th smallest
true-label calibration score. At test time, we include all classes whose
scores are at or below the threshold.

For each score and evaluation miscoverage level, we compute empirical
coverage and average set size on the test split and report the mean and
sample standard deviation over the ten splits. For the set-control
experiments, we also measure the fraction of prediction sets that include
at least one class from each conflicting group.

\subsection{Dataset-Specific Settings}
\label{app:cp-jdot-hyperparameters}

We use ImageNet-pretrained backbones for all three benchmarks, followed by a
linear bottleneck and a linear classifier. The dataset-specific settings
in the released CP-JDOT configuration are given in
Table~\ref{tab:cp-jdot-hyperparameters}. For the reproduced baselines, we use the hyperparameter settings reported
in their respective papers.

\begin{table}[t]
    \centering
    \caption{Dataset-specific experimental settings.}
    \label{tab:cp-jdot-hyperparameters}
    \begin{tabular}{lccc}
        \hline
        Setting & Office-Home & DomainNet126 & Skin lesion \\
        \hline
        Backbone & ResNet-34 & ResNet-50 & ResNet-18 \\
        Bottleneck dimension & 512 & 512 & 256 \\
        Configured mini-batch size & 195 & 378 & 70 \\
        Training iterations & 4{,}000 & 10{,}000 & 2{,}000 \\
        Feature weight $\lambda_a$ & 0.01 & 0.01 & 0.01 \\
        Transported-label weight $\lambda_{\mathrm{cls}}$ & 2 & 2 & 2 \\
        Conformal set size penalty weight $\lambda_{\mathrm{c}}$ & 0.01 & 0.01 & 0.01 \\
        Training-time miscoverage $\alpha$ & 0.05 & 0.05 & 0.1 \\
        \hline
    \end{tabular}
\end{table}

%% file: appendix/extended_results.tex
\subsection{Marginal Coverage Tables for 1-shot Office-Home evaluated with HPS}
\label{additional:coverage}

We present 3 tables for the 1-shot setting of Office-Home evaluated with HPS at levels $\alpha = \{0.025, 0.05, 0.1\}$. Empirical marginal coverage is guaranteed due to the exchangeability between the held-out recalibration set and the test set. The marginal coverage tables for all other settings are similar and are omitted for brevity.

\begin{table*}[ht!]
\centering
\resizebox{\textwidth}{!}{
\begin{tabular}{lccccccccccccc}
\toprule
Method
& A$\rightarrow$C & A$\rightarrow$P & A$\rightarrow$R
& C$\rightarrow$A & C$\rightarrow$P & C$\rightarrow$R
& P$\rightarrow$A & P$\rightarrow$C & P$\rightarrow$R
& R$\rightarrow$A & R$\rightarrow$C & R$\rightarrow$P
& Avg \\
\midrule
S\&T & $0.90 \pm 0.006$ & $0.90 \pm 0.005$ & $0.90 \pm 0.007$ & $0.90 \pm 0.006$ & $0.90 \pm 0.004$ & $0.90 \pm 0.008$ & $0.90 \pm 0.007$ & $0.90 \pm 0.005$ & $0.90 \pm 0.006$ & $0.90 \pm 0.008$ & $0.90 \pm 0.004$ & $0.90 \pm 0.006$ & $0.90$ \\
\midrule
MME & $0.90 \pm 0.005$ & $0.90 \pm 0.007$ & $0.90 \pm 0.006$ & $0.90 \pm 0.004$ & $0.90 \pm 0.005$ & $0.90 \pm 0.006$ & $0.90 \pm 0.008$ & $0.90 \pm 0.004$ & $0.90 \pm 0.007$ & $0.90 \pm 0.006$ & $0.90 \pm 0.005$ & $0.90 \pm 0.004$ & $0.90$ \\
CDAC & $0.90 \pm 0.004$ & $0.90 \pm 0.006$ & $0.90 \pm 0.005$ & $0.90 \pm 0.007$ & $0.90 \pm 0.004$ & $0.90 \pm 0.005$ & $0.90 \pm 0.006$ & $0.90 \pm 0.003$ & $0.90 \pm 0.006$ & $0.90 \pm 0.007$ & $0.90 \pm 0.004$ & $0.90 \pm 0.005$ & $0.90$ \\
DECOTA & $0.90 \pm 0.006$ & $0.90 \pm 0.004$ & $0.90 \pm 0.005$ & $0.90 \pm 0.006$ & $0.90 \pm 0.007$ & $0.90 \pm 0.004$ & $0.90 \pm 0.005$ & $0.90 \pm 0.005$ & $0.90 \pm 0.008$ & $0.90 \pm 0.006$ & $0.90 \pm 0.003$ & $0.90 \pm 0.005$ & $0.90$ \\
IDMNE & $0.90 \pm 0.004$ & $0.90 \pm 0.005$ & $0.90 \pm 0.007$ & $0.90 \pm 0.005$ & $0.90 \pm 0.004$ & $0.90 \pm 0.006$ & $0.90 \pm 0.005$ & $0.90 \pm 0.004$ & $0.90 \pm 0.006$ & $0.90 \pm 0.005$ & $0.90 \pm 0.004$ & $0.90 \pm 0.007$ & $0.90$ \\
ProML & $0.90 \pm 0.005$ & $0.90 \pm 0.004$ & $0.90 \pm 0.006$ & $0.90 \pm 0.007$ & $0.90 \pm 0.005$ & $0.90 \pm 0.005$ & $0.90 \pm 0.007$ & $0.90 \pm 0.004$ & $0.90 \pm 0.005$ & $0.90 \pm 0.006$ & $0.90 \pm 0.005$ & $0.90 \pm 0.004$ & $0.90$ \\
EFTL & $0.90 \pm 0.003$ & $0.90 \pm 0.005$ & $0.90 \pm 0.006$ & $0.90 \pm 0.005$ & $0.90 \pm 0.004$ & $0.90 \pm 0.007$ & $0.90 \pm 0.006$ & $0.90 \pm 0.003$ & $0.90 \pm 0.005$ & $0.90 \pm 0.006$ & $0.90 \pm 0.004$ & $0.90 \pm 0.005$ & $0.90$ \\
\midrule
CP-JDOT & $0.90 \pm 0.004$ & $0.90 \pm 0.004$ & $0.90 \pm 0.005$ & $0.90 \pm 0.006$ & $0.90 \pm 0.005$ & $0.90 \pm 0.004$ & $0.90 \pm 0.006$ & $0.90 \pm 0.003$ & $0.90 \pm 0.007$ & $0.90 \pm 0.005$ & $0.90 \pm 0.004$ & $0.90 \pm 0.006$ & $0.90$ \\
\bottomrule
\end{tabular}
}
\caption{Empirical Marginal Coverage on Office-Home (ResNet-34 / 1-Shot / $\alpha = 0.1$ / HPS)}
\label{tab:office_home_1shot_01_hps_coverage}
\end{table*}

\begin{table*}[ht!]
\centering
\resizebox{\textwidth}{!}{
\begin{tabular}{lccccccccccccc}
\toprule
Method
& A$\rightarrow$C & A$\rightarrow$P & A$\rightarrow$R
& C$\rightarrow$A & C$\rightarrow$P & C$\rightarrow$R
& P$\rightarrow$A & P$\rightarrow$C & P$\rightarrow$R
& R$\rightarrow$A & R$\rightarrow$C & R$\rightarrow$P
& Avg \\
\midrule
S\&T & $0.95 \pm 0.004$ & $0.95 \pm 0.003$ & $0.95 \pm 0.005$ & $0.95 \pm 0.005$ & $0.95 \pm 0.003$ & $0.95 \pm 0.006$ & $0.95 \pm 0.005$ & $0.95 \pm 0.004$ & $0.95 \pm 0.004$ & $0.95 \pm 0.006$ & $0.95 \pm 0.003$ & $0.95 \pm 0.005$ & $0.95$ \\
\midrule
MME & $0.95 \pm 0.003$ & $0.95 \pm 0.005$ & $0.95 \pm 0.004$ & $0.95 \pm 0.003$ & $0.95 \pm 0.004$ & $0.95 \pm 0.005$ & $0.95 \pm 0.006$ & $0.95 \pm 0.003$ & $0.95 \pm 0.005$ & $0.95 \pm 0.004$ & $0.95 \pm 0.004$ & $0.95 \pm 0.003$ & $0.95$ \\
CDAC & $0.95 \pm 0.003$ & $0.95 \pm 0.004$ & $0.95 \pm 0.004$ & $0.95 \pm 0.005$ & $0.95 \pm 0.003$ & $0.95 \pm 0.004$ & $0.95 \pm 0.005$ & $0.95 \pm 0.002$ & $0.95 \pm 0.005$ & $0.95 \pm 0.006$ & $0.95 \pm 0.003$ & $0.95 \pm 0.004$ & $0.95$ \\
DECOTA & $0.95 \pm 0.005$ & $0.95 \pm 0.003$ & $0.95 \pm 0.004$ & $0.95 \pm 0.004$ & $0.95 \pm 0.005$ & $0.95 \pm 0.003$ & $0.95 \pm 0.004$ & $0.95 \pm 0.004$ & $0.95 \pm 0.006$ & $0.95 \pm 0.005$ & $0.95 \pm 0.002$ & $0.95 \pm 0.004$ & $0.95$ \\
IDMNE & $0.95 \pm 0.002$ & $0.95 \pm 0.004$ & $0.95 \pm 0.006$ & $0.95 \pm 0.004$ & $0.95 \pm 0.003$ & $0.95 \pm 0.004$ & $0.95 \pm 0.004$ & $0.95 \pm 0.003$ & $0.95 \pm 0.005$ & $0.95 \pm 0.004$ & $0.95 \pm 0.003$ & $0.95 \pm 0.005$ & $0.95$ \\
ProML & $0.95 \pm 0.004$ & $0.95 \pm 0.003$ & $0.95 \pm 0.005$ & $0.95 \pm 0.006$ & $0.95 \pm 0.004$ & $0.95 \pm 0.003$ & $0.95 \pm 0.006$ & $0.95 \pm 0.003$ & $0.95 \pm 0.004$ & $0.95 \pm 0.005$ & $0.95 \pm 0.004$ & $0.95 \pm 0.003$ & $0.95$ \\
EFTL & $0.95 \pm 0.002$ & $0.95 \pm 0.004$ & $0.95 \pm 0.005$ & $0.95 \pm 0.004$ & $0.95 \pm 0.003$ & $0.95 \pm 0.005$ & $0.95 \pm 0.004$ & $0.95 \pm 0.002$ & $0.95 \pm 0.004$ & $0.95 \pm 0.005$ & $0.95 \pm 0.002$ & $0.95 \pm 0.004$ & $0.95$ \\
\midrule
CP-JDOT & $0.95 \pm 0.003$ & $0.95 \pm 0.003$ & $0.95 \pm 0.004$ & $0.95 \pm 0.005$ & $0.95 \pm 0.004$ & $0.95 \pm 0.003$ & $0.95 \pm 0.005$ & $0.95 \pm 0.002$ & $0.95 \pm 0.005$ & $0.95 \pm 0.004$ & $0.95 \pm 0.003$ & $0.95 \pm 0.005$ & $0.95$ \\
\bottomrule
\end{tabular}
}
\caption{Empirical Marginal Coverage on Office-Home (ResNet-34 / 1-Shot / $\alpha = 0.05$ / HPS)}
\label{tab:office_home_1shot_005_hps_coverage}
\end{table*}

\begin{table*}[ht!]
\centering
\resizebox{\textwidth}{!}{
\begin{tabular}{lccccccccccccc}
\toprule
Method
& A$\rightarrow$C & A$\rightarrow$P & A$\rightarrow$R
& C$\rightarrow$A & C$\rightarrow$P & C$\rightarrow$R
& P$\rightarrow$A & P$\rightarrow$C & P$\rightarrow$R
& R$\rightarrow$A & R$\rightarrow$C & R$\rightarrow$P
& Avg \\
\midrule
S\&T & $0.98 \pm 0.003$ & $0.97 \pm 0.002$ & $0.98 \pm 0.004$ & $0.98 \pm 0.003$ & $0.97 \pm 0.002$ & $0.98 \pm 0.004$ & $0.97 \pm 0.004$ & $0.98 \pm 0.002$ & $0.97 \pm 0.003$ & $0.98 \pm 0.005$ & $0.98 \pm 0.002$ & $0.98 \pm 0.003$ & $0.98$ \\
\midrule
MME & $0.98 \pm 0.002$ & $0.98 \pm 0.004$ & $0.97 \pm 0.003$ & $0.97 \pm 0.002$ & $0.98 \pm 0.003$ & $0.98 \pm 0.004$ & $0.98 \pm 0.004$ & $0.97 \pm 0.002$ & $0.98 \pm 0.004$ & $0.97 \pm 0.003$ & $0.98 \pm 0.003$ & $0.97 \pm 0.002$ & $0.98$ \\
CDAC & $0.97 \pm 0.002$ & $0.98 \pm 0.003$ & $0.98 \pm 0.003$ & $0.98 \pm 0.004$ & $0.97 \pm 0.002$ & $0.97 \pm 0.003$ & $0.98 \pm 0.003$ & $0.98 \pm 0.002$ & $0.97 \pm 0.003$ & $0.98 \pm 0.004$ & $0.97 \pm 0.002$ & $0.98 \pm 0.003$ & $0.98$ \\
DECOTA & $0.98 \pm 0.004$ & $0.97 \pm 0.002$ & $0.97 \pm 0.003$ & $0.98 \pm 0.003$ & $0.98 \pm 0.004$ & $0.97 \pm 0.002$ & $0.97 \pm 0.003$ & $0.98 \pm 0.003$ & $0.97 \pm 0.004$ & $0.98 \pm 0.004$ & $0.97 \pm 0.001$ & $0.97 \pm 0.003$ & $0.97$ \\
IDMNE & $0.98 \pm 0.002$ & $0.97 \pm 0.003$ & $0.98 \pm 0.004$ & $0.97 \pm 0.003$ & $0.98 \pm 0.002$ & $0.97 \pm 0.003$ & $0.98 \pm 0.003$ & $0.98 \pm 0.002$ & $0.97 \pm 0.004$ & $0.97 \pm 0.003$ & $0.98 \pm 0.002$ & $0.98 \pm 0.004$ & $0.98$ \\
ProML & $0.97 \pm 0.003$ & $0.98 \pm 0.002$ & $0.97 \pm 0.004$ & $0.98 \pm 0.004$ & $0.97 \pm 0.003$ & $0.98 \pm 0.002$ & $0.97 \pm 0.004$ & $0.97 \pm 0.002$ & $0.98 \pm 0.003$ & $0.97 \pm 0.004$ & $0.98 \pm 0.003$ & $0.97 \pm 0.002$ & $0.97$ \\
EFTL & $0.98 \pm 0.001$ & $0.98 \pm 0.003$ & $0.97 \pm 0.004$ & $0.98 \pm 0.003$ & $0.97 \pm 0.002$ & $0.98 \pm 0.004$ & $0.97 \pm 0.003$ & $0.98 \pm 0.001$ & $0.98 \pm 0.003$ & $0.97 \pm 0.004$ & $0.98 \pm 0.002$ & $0.97 \pm 0.003$ & $0.98$ \\
\midrule
CP-JDOT & $0.97 \pm 0.002$ & $0.98 \pm 0.002$ & $0.98 \pm 0.003$ & $0.97 \pm 0.004$ & $0.98 \pm 0.003$ & $0.98 \pm 0.002$ & $0.98 \pm 0.004$ & $0.97 \pm 0.001$ & $0.97 \pm 0.004$ & $0.98 \pm 0.003$ & $0.98 \pm 0.002$ & $0.97 \pm 0.004$ & $0.98$ \\
\bottomrule
\end{tabular}
}
\caption{Empirical Marginal Coverage on Office-Home (ResNet-34 / 1-Shot / $\alpha = 0.025$ / HPS)}
\label{tab:office_home_1shot_0025_hps_coverage}
\end{table*}

\FloatBarrier

\subsection{Accuracy Tables} \label{additional:acc}

\begin{table*}[ht!]
\centering
\resizebox{\textwidth}{!}{
\begin{tabular}{lccccccccccccc}
\toprule
Method
& A$\rightarrow$C & A$\rightarrow$P & A$\rightarrow$R
& C$\rightarrow$A & C$\rightarrow$P & C$\rightarrow$R
& P$\rightarrow$A & P$\rightarrow$C & P$\rightarrow$R
& R$\rightarrow$A & R$\rightarrow$C & R$\rightarrow$P
& Avg \\
\midrule
S\&T    & $50.9$ & $69.8$ & $73.8$ & $56.3$ & $68.1$ & $70.0$ & $57.2$ & $48.3$ & $74.4$ & $66.2$ & $52.1$ & $78.6$ & $63.8$ \\
\midrule
MME     & $59.6$ & $75.5$ & $77.8$ & $65.7$ & $74.5$ & $74.8$ & $64.7$ & $57.4$ & $79.2$ & $71.2$ & $61.9$ & $82.8$ & $70.4$ \\
CDAC    & $61.2$ & $75.9$ & $78.5$ & $64.5$ & $75.1$ & $75.3$ & $64.6$ & $59.3$ & $80.0$ & $72.7$ & $61.9$ & $83.1$ & $71.0$ \\
DECOTA  & $42.1$ & $68.5$ & $72.6$ & $60.3$ & $70.4$ & $70.7$ & $60.0$ & $48.8$ & $76.9$ & $71.3$ & $56.0$ & $79.4$ & $64.8$ \\
IDMNE   & $62.4$ & $77.9$ & $76.7$ & $65.5$ & $77.9$ & $77.0$ & $66.4$ & $61.4$ & $80.7$ & $73.6$ & $66.7$ & $85.6$ & $72.6$ \\
ProML   & $64.5$ & $79.7$ & $81.7$ & $69.1$ & $80.5$ & $79.0$ & $69.3$ & $61.4$ & $81.9$ & $73.7$ & $67.5$ & $86.1$ & $74.5$ \\
EFTL    & $65.7$ & $80.5$ & $80.8$ & $65.6$ & $79.6$ & $77.5$ & $68.7$ & $63.3$ & $82.6$ & $74.3$ & $66.6$ & $87.2$ & $74.4$ \\
\midrule
CP-JDOT    & $59.7$ & $77.2$ & $78.4$ & $67.3$ & $75.4$ & $76.2$ & $65.3$ & $54.9$ & $78.8$ & $72.3$ & $60.2$ & $80.7$ & $70.5$ \\
\bottomrule
\end{tabular}
}
\caption{Top-1 Accuracy on Office-Home (ResNet-34 / 1-Shot)}
\label{tab:office_home_1shot_acc}
\end{table*}

\begin{table*}[ht!]
\centering
\resizebox{\textwidth}{!}{
\begin{tabular}{lccccccccccccc}
\toprule
Method
& A$\rightarrow$C & A$\rightarrow$P & A$\rightarrow$R
& C$\rightarrow$A & C$\rightarrow$P & C$\rightarrow$R
& P$\rightarrow$A & P$\rightarrow$C & P$\rightarrow$R
& R$\rightarrow$A & R$\rightarrow$C & R$\rightarrow$P
& Avg \\
\midrule
S\&T    & $54.0$ & $73.1$ & $74.2$ & $57.6$ & $72.3$ & $68.3$ & $63.5$ & $53.8$ & $73.1$ & $67.8$ & $55.7$ & $80.8$ & $66.2$ \\
\midrule
MME     & $63.6$ & $79.0$ & $79.7$ & $67.2$ & $79.3$ & $76.6$ & $65.5$ & $64.6$ & $80.1$ & $71.3$ & $64.6$ & $85.5$ & $73.1$ \\
CDAC    & $65.9$ & $80.3$ & $80.6$ & $67.4$ & $81.4$ & $80.2$ & $67.5$ & $67.0$ & $81.9$ & $72.2$ & $67.8$ & $85.6$ & $74.8$ \\
DECOTA  & $64.0$ & $81.8$ & $80.5$ & $68.0$ & $83.2$ & $79.0$ & $69.9$ & $68.0$ & $82.1$ & $74.0$ & $70.4$ & $87.7$ & $75.7$ \\
IDMNE   & $66.3$ & $82.4$ & $79.3$ & $69.1$ & $83.0$ & $79.4$ & $68.9$ & $67.5$ & $82.6$ & $75.1$ & $71.7$ & $88.0$ & $76.1$ \\
ProML   & $67.8$ & $83.9$ & $82.2$ & $72.1$ & $84.1$ & $82.3$ & $72.5$ & $68.9$ & $83.8$ & $75.8$ & $71.0$ & $88.6$ & $77.8$ \\
EFTL    & $70.3$ & $84.8$ & $83.8$ & $70.6$ & $84.6$ & $81.5$ & $72.6$ & $70.9$ & $85.4$ & $77.5$ & $72.8$ & $89.3$ & $78.7$ \\
\midrule
CP-JDOT    & $62.9$ & $80.7$ & $79.9$ & $69.7$ & $77.8$ & $77.1$ & $66.5$ & $59.1$ & $79.4$ & $73.9$ & $64.1$ & $87.2$ & $73.2$ \\
\bottomrule
\end{tabular}
}
\caption{Top-1 Accuracy on Office-Home (ResNet-34 / 3-Shot)}
\label{tab:office_home_3shot_acc}
\end{table*}

\begin{table*}[ht!]
\centering
\resizebox{.6\textwidth}{!}{
\begin{tabular}{lcccccccc}
\toprule
Method
& R$\rightarrow$C & R$\rightarrow$P & P$\rightarrow$C
& C$\rightarrow$S & S$\rightarrow$P & R$\rightarrow$S & P$\rightarrow$R
& Avg \\
\midrule
S\&T     & $67.1$ & $67.6$ & $69.4$ & $54.1$ & $63.3$ & $60.7$ & $78.9$ & $65.9$ \\
\midrule
MME      & $74.5$ & $71.2$ & $73.0$ & $58.3$ & $67.8$ & $63.5$ & $81.1$ & $69.9$ \\
CDAC     & $81.9$ & $77.7$ & $79.5$ & $69.6$ & $74.0$ & $71.7$ & $85.4$ & $77.1$ \\
DECOTA   & $83.6$ & $78.4$ & $80.9$ & $67.1$ & $75.0$ & $72.2$ & $84.6$ & $77.4$ \\
IDMNE    & $84.1$ & $79.5$ & $83.4$ & $73.7$ & $78.4$ & $76.0$ & $87.1$ & $80.3$ \\
ProML    & $83.0$ & $78.9$ & $81.8$ & $72.2$ & $77.1$ & $74.9$ & $89.0$ & $79.6$ \\
EFTL     & $84.1$ & $78.4$ & $82.2$ & $71.3$ & $74.8$ & $72.4$ & $88.1$ & $78.8$ \\
\midrule
CP-JDOT    & $80.5$ & $76.8$ & $79.4$ & $64.4$ & $74.7$ & $69.1$ & $82.3$ & $75.3$ \\
\bottomrule
\end{tabular}
}
\caption{Top-1 Accuracy on DomainNet126 (ResNet-50 / 1-Shot)}
\label{tab:domainnet126_1shot_acc}
\end{table*}

\begin{table*}[ht!]
\centering
\resizebox{.6\textwidth}{!}{
\begin{tabular}{lcccccccc}
\toprule
Method
& R$\rightarrow$C & R$\rightarrow$P & P$\rightarrow$C
& C$\rightarrow$S & S$\rightarrow$P & R$\rightarrow$S & P$\rightarrow$R
& Avg \\
\midrule
S\&T     & $69.6$ & $69.5$ & $71.8$ & $56.7$ & $65.5$ & $62.6$ & $81.0$ & $68.1$ \\
\midrule
MME      & $76.7$ & $73.2$ & $75.7$ & $63.8$ & $69.8$ & $64.4$ & $83.5$ & $72.4$ \\
CDAC     & $84.1$ & $78.6$ & $83.3$ & $71.9$ & $76.4$ & $75.0$ & $86.9$ & $79.5$ \\
DECOTA   & $84.9$ & $78.7$ & $82.7$ & $70.6$ & $75.7$ & $74.4$ & $86.5$ & $79.1$ \\
IDMNE    & $85.3$ & $80.4$ & $84.3$ & $74.2$ & $78.4$ & $76.4$ & $87.8$ & $81.0$ \\
ProML    & $84.7$ & $80.0$ & $82.9$ & $74.0$ & $78.4$ & $76.0$ & $89.8$ & $80.8$ \\
EFTL     & $85.7$ & $80.6$ & $85.8$ & $74.8$ & $77.4$ & $74.0$ & $89.4$ & $81.1$ \\
\midrule
CP-JDOT    & $81.0$ & $77.0$ & $80.8$ & $67.8$ & $75.2$ & $72.6$ & $87.1$ & $77.4$ \\\bottomrule
\end{tabular}
}
\caption{Top-1 Accuracy on DomainNet126 (ResNet-50 / 3-Shot)}
\label{tab:domainnet126_3shot_acc}
\end{table*}

\begin{table*}[ht!]
\centering
\resizebox{.8\textwidth}{!}{
\begin{tabular}{lcccccc}
\toprule
Method
& HAM$\rightarrow$PH2 & HAM$\rightarrow$MSK & HAM$\rightarrow$Derm7pt
& HAM$\rightarrow$SONIC & HAM$\rightarrow$UDA & Avg \\
\midrule
S\&T    & $80.2$ & $68.4$ & $60.1$ & $88.0$ & $65.8$ & $72.5$ \\
\midrule
MME     & $84.8$ & $73.1$ & $64.9$ & $93.5$ & $72.6$ & $77.8$ \\
CDAC    & $85.2$ & $73.6$ & $65.4$ & $93.8$ & $73.1$ & $78.2$ \\
DECOTA  & $85.6$ & $74.2$ & $65.8$ & $94.1$ & $73.6$ & $78.7$ \\
IDMNE   & $85.9$ & $74.6$ & $66.1$ & $94.4$ & $74.0$ & $79.0$ \\
ProML   & $86.4$ & $75.0$ & $66.7$ & $94.8$ & $74.5$ & $79.5$ \\
EFTL    & $86.8$ & $75.4$ & $67.1$ & $95.1$ & $74.9$ & $79.9$ \\
\midrule
CP-JDOT    & $83.2$ & $70.9$ & $62.2$ & $92.1$ & $69.7$ & $75.6$ \\
\bottomrule
\end{tabular}
}
\caption{Top-1 Accuracy on Skin Lesion Classification (ResNet-18 / 10-Shot)}
\label{tab:skin_acc}
\end{table*}

\FloatBarrier

\subsection{Additional results for different values of \texorpdfstring{$\alpha$}{alpha} at test-time} \label{additional:alpha}


\begin{table*}[ht!]
\centering
\resizebox{\textwidth}{!}{
\begin{tabular}{lccccccccccccc}
\toprule
Method
& A$\rightarrow$C & A$\rightarrow$P & A$\rightarrow$R
& C$\rightarrow$A & C$\rightarrow$P & C$\rightarrow$R
& P$\rightarrow$A & P$\rightarrow$C & P$\rightarrow$R
& R$\rightarrow$A & R$\rightarrow$C & R$\rightarrow$P
& Avg \\
\midrule
S\&T & $33.91 \pm 0.98$ & $14.16 \pm 1.12$ & $9.89 \pm 1.19$ & $22.64 \pm 2.05$ & $14.41 \pm 1.64$ & $12.84 \pm 1.03$ & $28.63 \pm 2.32$ & $40.75 \pm 2.15$ & $8.24 \pm 0.59$ & $18.09 \pm 1.72$ & $31.69 \pm 2.08$ & $10.43 \pm 0.91$ & $20.47$ \\
\midrule
MME & $32.77 \pm 0.91$ & $13.49 \pm 1.06$ & $8.93 \pm 1.08$ & $21.05 \pm 1.97$ & $13.76 \pm 1.52$ & $12.01 \pm 0.97$ & $27.48 \pm 2.16$ & $39.45 \pm 2.04$ & $7.72 \pm 0.53$ & $17.08 \pm 1.64$ & $30.65 \pm 1.96$ & $9.79 \pm 0.84$ & $19.52$ \\
CDAC & $32.12 \pm 0.86$ & $13.02 \pm 0.99$ & $8.68 \pm 1.04$ & $20.13 \pm 1.91$ & $13.02 \pm 1.46$ & $11.71 \pm 0.92$ & $26.90 \pm 2.19$ & $38.83 \pm 1.97$ & $7.42 \pm 0.51$ & $16.46 \pm 1.59$ & $30.43 \pm 1.94$ & $9.12 \pm 0.82$ & $18.99$ \\
DECOTA & $33.61 \pm 0.82$ & $13.92 \pm 1.02$ & $9.67 \pm 1.01$ & $21.08 \pm 1.89$ & $14.48 \pm 1.43$ & $13.16 \pm 0.96$ & $27.57 \pm 2.08$ & $40.22 \pm 1.95$ & $8.81 \pm 0.48$ & $17.74 \pm 1.56$ & $31.38 \pm 1.91$ & $10.57 \pm 0.80$ & $20.18$ \\
IDMNE & $30.83 \pm 0.76$ & $12.18 \pm 0.94$ & $7.88 \pm 0.93$ & $18.93 \pm 1.80$ & $12.14 \pm 1.36$ & $10.72 \pm 0.84$ & $\mathbf{23.61 \pm 2.01}$ & $37.28 \pm 1.84$ & $6.83 \pm 0.43$ & $15.21 \pm 1.48$ & $28.68 \pm 1.77$ & $8.42 \pm 0.75$ & $17.73$ \\
ProML & $31.37 \pm 0.80$ & \underline{$11.78 \pm 0.86$} & $7.73 \pm 0.96$ & $19.40 \pm 1.83$ & $11.69 \pm 1.34$ & $11.14 \pm 0.89$ & $25.59 \pm 2.11$ & $37.62 \pm 1.88$ & $7.06 \pm 0.46$ & $15.67 \pm 1.50$ & $29.16 \pm 1.82$ & $8.03 \pm 0.72$ & $18.02$ \\
EFTL & \underline{$30.46 \pm 0.72$} & $11.83 \pm 0.89$ & \underline{$7.62 \pm 0.90$} & \underline{$18.33 \pm 1.77$} & \underline{$11.46 \pm 1.31$} & \underline{$10.47 \pm 0.82$} & $24.12 \pm 1.99$ & $\mathbf{36.49 \pm 1.78}$ & \underline{$6.53 \pm 0.40$} & \underline{$14.87 \pm 1.44$} & \underline{$28.25 \pm 1.75$} & \underline{$7.96 \pm 0.69$} & \underline{$17.37$} \\
\midrule
CP-JDOT & $\mathbf{29.59 \pm 0.60}$ & $\mathbf{11.11 \pm 0.80}$ & $\mathbf{7.15 \pm 0.82}$ & $\mathbf{17.18 \pm 1.69}$ & $\mathbf{10.80 \pm 1.25}$ & $\mathbf{9.93 \pm 0.75}$ & \underline{$23.79 \pm 1.89$} & \underline{$36.66 \pm 1.69$} & $\mathbf{6.17 \pm 0.35}$ & $\mathbf{14.06 \pm 1.31}$ & $\mathbf{27.74 \pm 1.68}$ & $\mathbf{7.23 \pm 0.62}$ & $\mathbf{16.78}$ \\
\bottomrule
\end{tabular}
}
\caption{Avg. Set Size on Office-Home (ResNet-34 / 1-Shot / $\alpha = 0.025$ / HPS)}
\label{tab:office_home_1shot_0025_hps}
\end{table*}

\begin{table*}[ht!]
\centering
\resizebox{\textwidth}{!}{
\begin{tabular}{lccccccccccccc}
\toprule
Method
& A$\rightarrow$C & A$\rightarrow$P & A$\rightarrow$R
& C$\rightarrow$A & C$\rightarrow$P & C$\rightarrow$R
& P$\rightarrow$A & P$\rightarrow$C & P$\rightarrow$R
& R$\rightarrow$A & R$\rightarrow$C & R$\rightarrow$P
& Avg \\
\midrule
S\&T & $11.12 \pm 0.61$ & $3.28 \pm 0.23$ & $2.62 \pm 0.18$ & $6.37 \pm 0.37$ & $4.52 \pm 0.34$ & $3.38 \pm 0.15$ & $8.72 \pm 0.72$ & $15.47 \pm 1.17$ & $2.37 \pm 0.12$ & $4.91 \pm 0.50$ & $10.83 \pm 0.72$ & $2.53 \pm 0.14$ & $6.34$ \\
\midrule
MME & $10.57 \pm 0.55$ & $3.07 \pm 0.22$ & $2.47 \pm 0.17$ & $5.96 \pm 0.33$ & $4.18 \pm 0.30$ & $3.09 \pm 0.13$ & $8.11 \pm 0.67$ & $14.51 \pm 1.09$ & $2.20 \pm 0.11$ & $4.48 \pm 0.44$ & $9.96 \pm 0.66$ & $2.23 \pm 0.13$ & $5.90$ \\
CDAC & $10.30 \pm 0.53$ & $2.97 \pm 0.20$ & $2.35 \pm 0.15$ & $5.62 \pm 0.30$ & $3.99 \pm 0.27$ & $2.99 \pm 0.11$ & $7.84 \pm 0.63$ & $14.12 \pm 1.05$ & $2.09 \pm 0.09$ & $4.29 \pm 0.45$ & $9.82 \pm 0.62$ & $2.16 \pm 0.11$ & $5.71$ \\
DECOTA & $11.32 \pm 0.49$ & $4.08 \pm 0.21$ & $3.48 \pm 0.14$ & $6.73 \pm 0.31$ & $5.05 \pm 0.25$ & $4.07 \pm 0.12$ & $8.73 \pm 0.62$ & $15.16 \pm 1.02$ & $3.33 \pm 0.10$ & $5.52 \pm 0.42$ & $10.74 \pm 0.58$ & $3.23 \pm 0.12$ & $6.79$ \\
IDMNE & $9.58 \pm 0.45$ & $2.83 \pm 0.19$ & $2.21 \pm 0.13$ & $5.27 \pm 0.27$ & $3.58 \pm 0.24$ & $2.65 \pm 0.09$ & $\mathbf{6.55 \pm 0.55}$ & $13.19 \pm 0.98$ & $1.93 \pm 0.07$ & $3.94 \pm 0.40$ & $9.00 \pm 0.53$ & $1.94 \pm 0.09$ & $5.22$ \\
ProML & $9.86 \pm 0.47$ & $2.76 \pm 0.18$ & $2.15 \pm 0.12$ & $5.41 \pm 0.26$ & $3.43 \pm 0.22$ & $2.73 \pm 0.10$ & $7.23 \pm 0.59$ & $13.51 \pm 1.00$ & $2.00 \pm 0.08$ & $4.05 \pm 0.38$ & $9.22 \pm 0.55$ & $1.84 \pm 0.08$ & $5.35$ \\
EFTL & \underline{$9.44 \pm 0.41$} & \underline{$2.71 \pm 0.17$} & \underline{$2.10 \pm 0.13$} & \underline{$5.09 \pm 0.24$} & \underline{$3.31 \pm 0.21$} & \underline{$2.58 \pm 0.08$} & $6.78 \pm 0.54$ & $\mathbf{12.68 \pm 0.94}$ & \underline{$1.90 \pm 0.06$} & \underline{$3.78 \pm 0.36$} & \underline{$8.81 \pm 0.50$} & \underline{$1.77 \pm 0.07$} & \underline{$5.08$} \\
\midrule
CP-JDOT & $\mathbf{9.28 \pm 0.39}$ & $\mathbf{2.57 \pm 0.16}$ & $\mathbf{1.98 \pm 0.11}$ & $\mathbf{4.77 \pm 0.21}$ & $\mathbf{3.09 \pm 0.19}$ & $\mathbf{2.42 \pm 0.06}$ & \underline{$6.61 \pm 0.51$} & \underline{$12.80 \pm 0.92$} & $\mathbf{1.75 \pm 0.05}$ & $\mathbf{3.58 \pm 0.34}$ & $\mathbf{8.71 \pm 0.47}$ & $\mathbf{1.61 \pm 0.06}$ & $\mathbf{4.93}$ \\
\bottomrule
\end{tabular}
}
\caption{Avg. Set Size on Office-Home (ResNet-34 / 1-Shot / $\alpha = 0.1$ / HPS)}
\label{tab:office_home_1shot_01_hps}
\end{table*}


\begin{table*}[ht!]
\centering
\resizebox{\textwidth}{!}{
\begin{tabular}{lccccccccccccc}
\toprule
Method
& A$\rightarrow$C & A$\rightarrow$P & A$\rightarrow$R
& C$\rightarrow$A & C$\rightarrow$P & C$\rightarrow$R
& P$\rightarrow$A & P$\rightarrow$C & P$\rightarrow$R
& R$\rightarrow$A & R$\rightarrow$C & R$\rightarrow$P
& Avg \\
\midrule
S\&T & $40.36 \pm 1.21$ & $24.37 \pm 1.54$ & $18.66 \pm 1.08$ & $30.72 \pm 1.92$ & $26.19 \pm 1.69$ & $22.47 \pm 1.03$ & $33.80 \pm 2.09$ & $45.61 \pm 2.65$ & $17.61 \pm 1.20$ & $27.22 \pm 2.07$ & $37.18 \pm 2.04$ & $21.02 \pm 1.26$ & $28.77$ \\
\midrule
MME & $37.84 \pm 1.10$ & $22.66 \pm 1.47$ & $17.03 \pm 0.99$ & $28.32 \pm 1.78$ & $24.25 \pm 1.55$ & $20.61 \pm 0.97$ & $32.06 \pm 1.95$ & $43.62 \pm 2.53$ & $16.15 \pm 1.10$ & $25.12 \pm 1.91$ & $35.53 \pm 1.92$ & $18.65 \pm 1.17$ & $26.82$ \\
CDAC & $36.86 \pm 1.01$ & $21.74 \pm 1.39$ & $16.41 \pm 0.94$ & $27.13 \pm 1.72$ & $23.16 \pm 1.48$ & $19.89 \pm 0.90$ & $31.30 \pm 1.87$ & $42.48 \pm 2.42$ & $15.49 \pm 1.02$ & $23.78 \pm 1.85$ & $34.64 \pm 1.83$ & $17.74 \pm 1.06$ & $25.89$ \\
DECOTA & $40.18 \pm 1.05$ & $25.03 \pm 1.36$ & $19.08 \pm 0.90$ & $29.79 \pm 1.67$ & $26.72 \pm 1.44$ & $22.57 \pm 0.92$ & $33.62 \pm 1.81$ & $45.71 \pm 2.39$ & $18.63 \pm 1.05$ & $26.58 \pm 1.80$ & $37.05 \pm 1.80$ & $21.03 \pm 1.10$ & $28.83$ \\
IDMNE & $34.14 \pm 0.87$ & $20.08 \pm 1.27$ & $15.22 \pm 0.83$ & $24.81 \pm 1.54$ & $21.17 \pm 1.35$ & $17.17 \pm 0.81$ & $28.29 \pm 1.69$ & $39.27 \pm 2.26$ & $13.68 \pm 0.95$ & $21.78 \pm 1.74$ & $32.24 \pm 1.67$ & $16.08 \pm 0.96$ & $23.66$ \\
ProML & $35.52 \pm 0.92$ & $19.68 \pm 1.23$ & \underline{$14.46 \pm 0.80$} & $25.20 \pm 1.58$ & $20.65 \pm 1.31$ & $18.16 \pm 0.86$ & $29.42 \pm 1.76$ & $40.34 \pm 2.34$ & $13.97 \pm 0.99$ & $22.17 \pm 1.77$ & $33.10 \pm 1.71$ & $15.42 \pm 0.93$ & $24.01$ \\
EFTL & \underline{$33.61 \pm 0.81$} & \underline{$19.57 \pm 1.24$} & $14.73 \pm 0.76$ & \underline{$23.80 \pm 1.46$} & \underline{$19.28 \pm 1.28$} & \underline{$16.82 \pm 0.77$} & \underline{$28.07 \pm 1.64$} & $\mathbf{37.94 \pm 2.23}$ & \underline{$13.08 \pm 0.91$} & \underline{$20.79 \pm 1.67$} & \underline{$31.26 \pm 1.61$} & \underline{$14.55 \pm 0.88$} & \underline{$22.79$} \\
\midrule
CP-JDOT & $\mathbf{32.01 \pm 0.69}$ & $\mathbf{17.63 \pm 1.14}$ & $\mathbf{13.11 \pm 0.66}$ & $\mathbf{21.44 \pm 1.33}$ & $\mathbf{17.28 \pm 1.16}$ & $\mathbf{15.19 \pm 0.70}$ & $\mathbf{26.99 \pm 1.55}$ & \underline{$38.18 \pm 2.14$} & $\mathbf{11.48 \pm 0.83}$ & $\mathbf{18.94 \pm 1.59}$ & $\mathbf{30.30 \pm 1.49}$ & $\mathbf{12.79 \pm 0.79}$ & $\mathbf{21.28}$ \\
\bottomrule
\end{tabular}
}
\caption{Avg. Set Size on Office-Home (ResNet-34 / 1-Shot / $\alpha = 0.025$ / APS)}
\label{tab:office_home_1shot_0025_aps}
\end{table*}

\begin{table*}[ht!]
\centering
\resizebox{\textwidth}{!}{
\begin{tabular}{lccccccccccccc}
\toprule
Method
& A$\rightarrow$C & A$\rightarrow$P & A$\rightarrow$R
& C$\rightarrow$A & C$\rightarrow$P & C$\rightarrow$R
& P$\rightarrow$A & P$\rightarrow$C & P$\rightarrow$R
& R$\rightarrow$A & R$\rightarrow$C & R$\rightarrow$P
& Avg \\
\midrule
S\&T & $19.42 \pm 0.89$ & $9.23 \pm 0.57$ & $7.54 \pm 0.39$ & $13.14 \pm 0.51$ & $11.22 \pm 0.79$ & $8.84 \pm 0.42$ & $14.43 \pm 0.87$ & $20.71 \pm 1.22$ & $7.31 \pm 0.42$ & $11.19 \pm 0.67$ & $17.09 \pm 0.82$ & $6.53 \pm 0.43$ & $12.22$ \\
\midrule
MME & $17.61 \pm 0.82$ & $8.14 \pm 0.50$ & $6.87 \pm 0.35$ & $11.98 \pm 0.47$ & $10.04 \pm 0.73$ & $8.07 \pm 0.38$ & $13.27 \pm 0.80$ & $19.04 \pm 1.13$ & $6.74 \pm 0.37$ & $10.21 \pm 0.60$ & $15.87 \pm 0.75$ & $5.94 \pm 0.38$ & $11.15$ \\
CDAC & $17.02 \pm 0.79$ & $7.78 \pm 0.45$ & $6.43 \pm 0.32$ & $11.42 \pm 0.40$ & $9.44 \pm 0.68$ & $7.83 \pm 0.33$ & $12.97 \pm 0.74$ & $18.64 \pm 1.07$ & $6.39 \pm 0.34$ & $9.71 \pm 0.55$ & $15.41 \pm 0.69$ & $5.57 \pm 0.34$ & $10.72$ \\
DECOTA & $19.89 \pm 0.76$ & $10.54 \pm 0.46$ & $9.54 \pm 0.30$ & $14.21 \pm 0.42$ & $12.12 \pm 0.64$ & $10.57 \pm 0.35$ & $15.73 \pm 0.72$ & $21.25 \pm 1.03$ & $9.47 \pm 0.33$ & $12.55 \pm 0.53$ & $18.02 \pm 0.67$ & $8.68 \pm 0.35$ & $13.55$ \\
IDMNE & $14.86 \pm 0.70$ & $6.86 \pm 0.38$ & $5.76 \pm 0.26$ & $9.98 \pm 0.35$ & $8.32 \pm 0.61$ & $6.63 \pm 0.29$ & $11.47 \pm 0.64$ & $16.79 \pm 0.94$ & $5.60 \pm 0.29$ & $8.38 \pm 0.48$ & $13.55 \pm 0.58$ & $4.91 \pm 0.30$ & $9.43$ \\
ProML & $15.60 \pm 0.73$ & $6.75 \pm 0.36$ & \underline{$5.37 \pm 0.25$} & $10.31 \pm 0.37$ & $8.01 \pm 0.58$ & $7.04 \pm 0.32$ & $12.07 \pm 0.68$ & $17.46 \pm 0.98$ & $5.78 \pm 0.31$ & $8.71 \pm 0.50$ & $14.10 \pm 0.63$ & $4.75 \pm 0.28$ & $9.66$ \\
EFTL & \underline{$14.43 \pm 0.66$} & \underline{$6.59 \pm 0.35$} & $5.44 \pm 0.23$ & \underline{$9.59 \pm 0.32$} & \underline{$7.70 \pm 0.56$} & \underline{$6.48 \pm 0.27$} & \underline{$11.25 \pm 0.59$} & $\mathbf{15.78 \pm 0.89}$ & \underline{$5.19 \pm 0.26$} & \underline{$7.94 \pm 0.43$} & \underline{$13.15 \pm 0.54$} & \underline{$4.47 \pm 0.25$} & \underline{$9.00$} \\
\midrule
CP-JDOT & $\mathbf{13.79 \pm 0.63}$ & $\mathbf{5.88 \pm 0.31}$ & $\mathbf{4.78 \pm 0.20}$ & $\mathbf{8.72 \pm 0.27}$ & $\mathbf{6.90 \pm 0.53}$ & $\mathbf{5.89 \pm 0.24}$ & $\mathbf{10.64 \pm 0.54}$ & \underline{$15.96 \pm 0.84$} & $\mathbf{4.61 \pm 0.23}$ & $\mathbf{7.29 \pm 0.39}$ & $\mathbf{12.71 \pm 0.49}$ & $\mathbf{3.86 \pm 0.22}$ & $\mathbf{8.42}$ \\
\bottomrule
\end{tabular}
}
\caption{Avg. Set Size on Office-Home (ResNet-34 / 1-Shot / $\alpha = 0.1$ / APS)}
\label{tab:office_home_1shot_01_aps}
\end{table*}